\documentclass{article}

\PassOptionsToPackage{numbers,sort&compress}{natbib}
\usepackage[preprint]{neurips_2026}

\usepackage[utf8]{inputenc}
\usepackage[T1]{fontenc}
\usepackage{hyperref}
\usepackage{url}
\usepackage{graphicx}
\usepackage{afterpage}
\usepackage{caption}
\usepackage{booktabs}
\usepackage{tabularx}
\usepackage{multirow}
\usepackage{amsfonts}
\usepackage{amsmath}
\usepackage{nicefrac}
\usepackage{microtype}
\usepackage[table]{xcolor}
\usepackage{listings}

\lstdefinestyle{promptstyle}{%
  basicstyle=\ttfamily\small,
  columns=fullflexible,
  breaklines=true,
  breakatwhitespace=false,
  breakindent=1.2em,
  keepspaces=true,
  showstringspaces=false,
  upquote=true,
  frame=single,
  framerule=0.3pt,
  rulecolor=\color{black!25},
  backgroundcolor=\color{black!3},
  framesep=4pt,
  framexleftmargin=4pt,
  framexrightmargin=4pt,
  framextopmargin=3pt,
  framexbottommargin=3pt,
  aboveskip=0.7em,
  belowskip=0.8em,
  xleftmargin=0.3em,
  xrightmargin=0.3em
}

\lstnewenvironment{PromptBox}{%
  \vspace{0.3em}%
  \lstset{style=promptstyle}%
}{}

\makeatletter
\let\oldthebibliography\thebibliography%
\renewcommand{\thebibliography}[1]{%
  \oldthebibliography{#1}%
  \small
}
\makeatother

\newcommand{\ModelCell}[2]{\rotatebox[origin=c]{90}{\shortstack[c]{\textbf{#1}\\\textbf{#2}}}}

\newcommand{\EmpiricalResultsPageEightTable}{%
\begin{table}[!t]
\centering
\small
\captionsetup{labelfont=bf}
\caption{\textbf{Counterfactual ablation} on LoCoMO with GPT-4o-mini. All counterfactual variants use \(k=15\) and a 32K offline context budget without rerunning answer generation. ``Unsup. Risk'' is an unsupported-evidence risk proxy.}
\label{tab:offline_ablation_main}
\setlength{\tabcolsep}{4pt}
\makebox[\linewidth][c]{%
\resizebox{1.0\linewidth}{!}{%
\begin{tabular}{l l c c c c c}
\toprule
Setting & Removed / Restricted Component
& Cand. Univ.
& Gold Ref Cov.
& Gold Traj. R@15
& Ctx. Tokens
& Unsup. Risk \\
\midrule
Full TrajWiki
& None
& 55.6
& 0.610
& 0.610
& 2.7K
& 0.645 \\

Direct Trajectory Retrieval
& Memory Wiki routing
& 130.4
& 0.356
& 0.346
& 25.6K
& 0.872 \\

Wiki Summaries Only
& Source-grounded trajectory expansion
& 41.2
& 0.000
& 0.807
& 25.3K
& 1.000 \\

Flat Raw-Memory Retrieval
& Trajectory and wiki organization
& 594.7
& 0.237
& 0.457
& 0.7K
& 0.918 \\

Latest Snapshot Only
& Historical snapshot depth
& 15.0
& 0.613
& 0.610
& 6.4K
& 0.649 \\
\bottomrule
\end{tabular}%
}%
}
\end{table}
}

\newcommand{\EmpiricalResultsPageEightFloats}{%
\noindent{\centering
\captionsetup{labelfont=bf,hypcap=false}
\begin{minipage}[t]{0.495\linewidth}
  \centering
  \includegraphics[width=\linewidth]{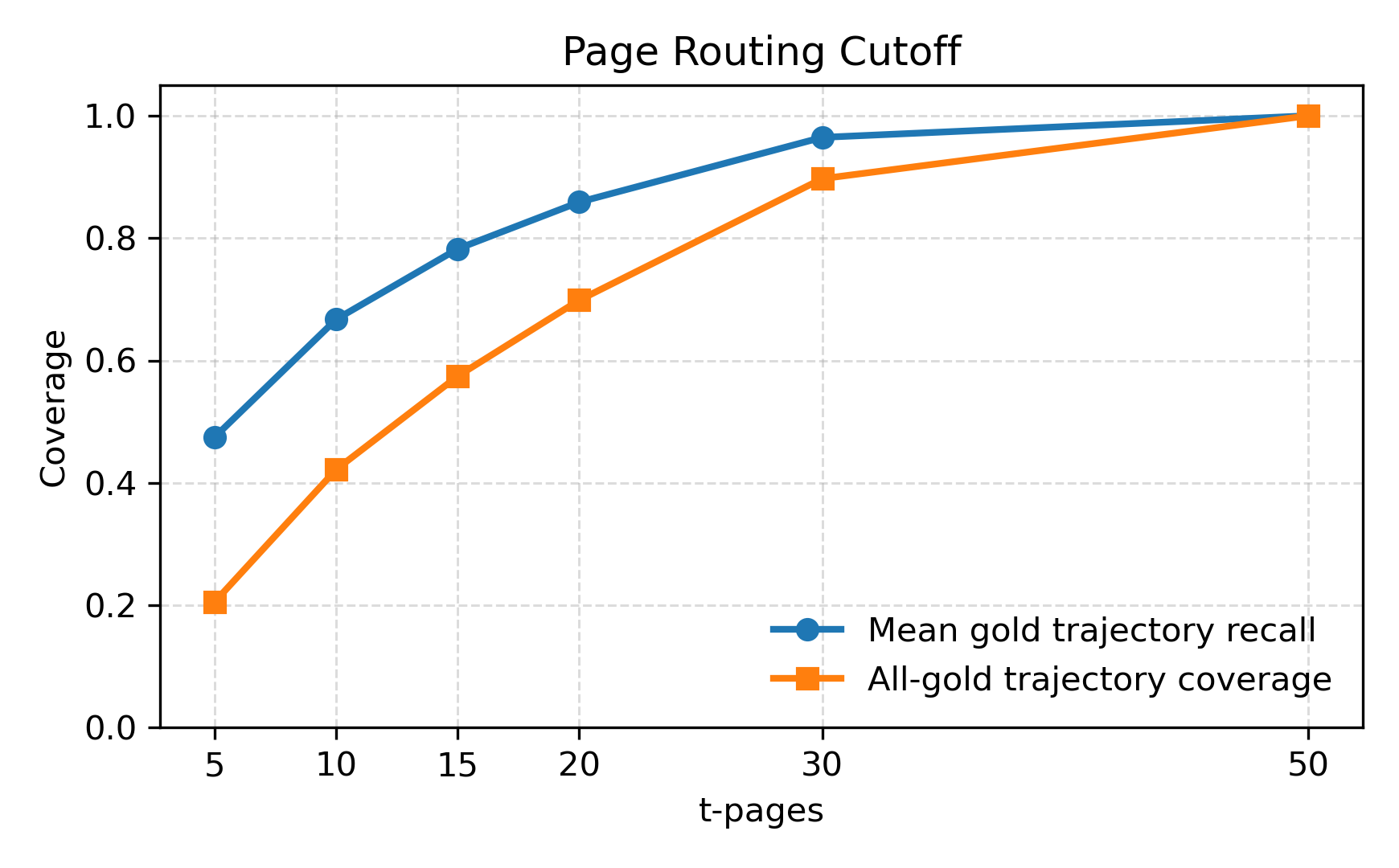}
\end{minipage}\hfill
\begin{minipage}[t]{0.495\linewidth}
  \centering
  \includegraphics[width=\linewidth]{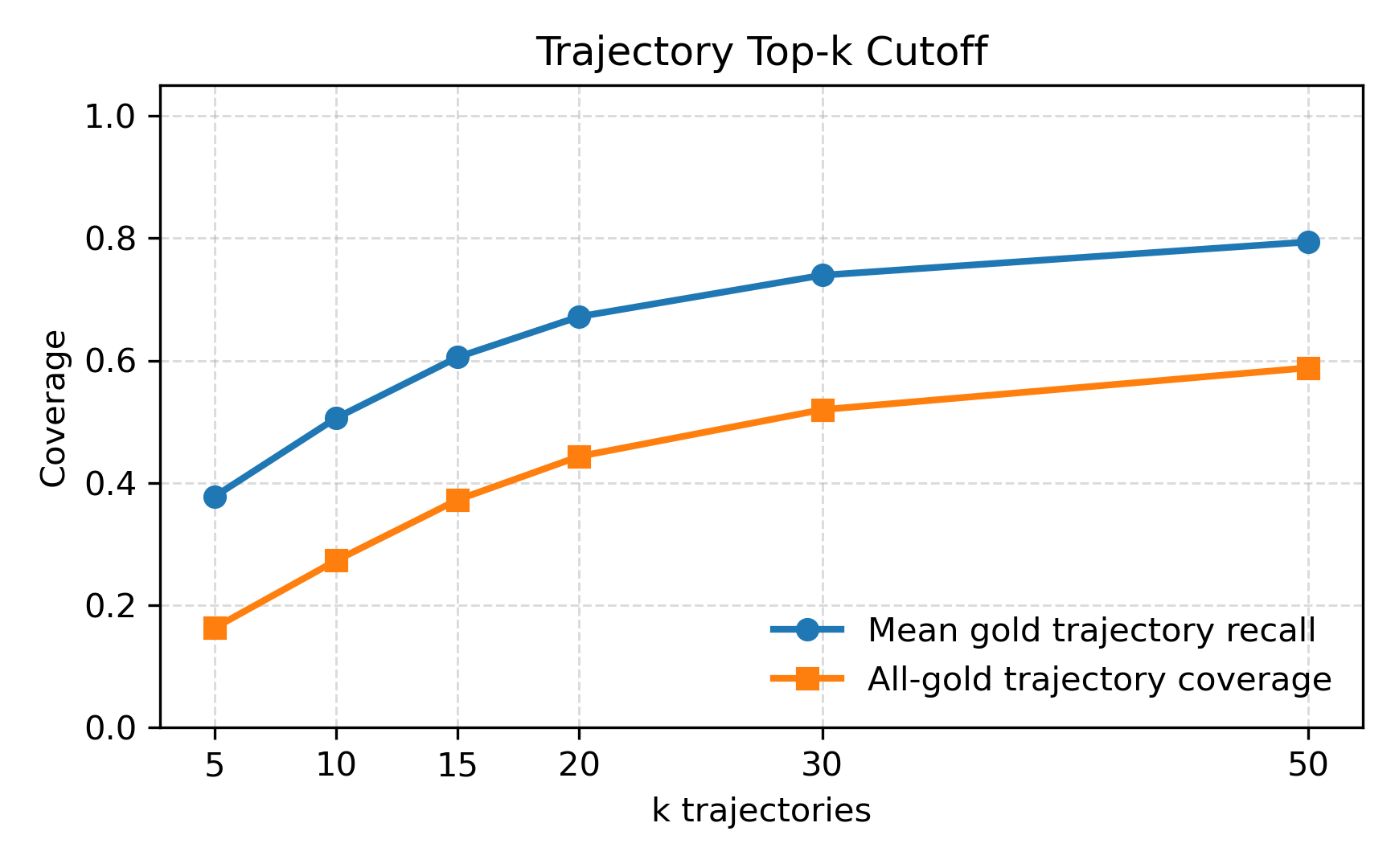}
\end{minipage}
\captionof{figure}{
\textbf{Hyperparameter Retrieval Cutoff Analysis} on the LoCoMo. 
The left plot varies the wiki page routing cutoff \(t\), showing how increasing routed pages improves gold trajectory coverage in the page-induced candidate universe. 
The right plot varies trajectory top-\(k\), showing that selecting more trajectories improves recall but increases the evidence passed to answer generation.
}
\label{fig:retrieval_cutoffs}
\par
\vspace{0.35em}
\begin{minipage}[t]{0.495\linewidth}
  \centering
  \includegraphics[width=\linewidth]{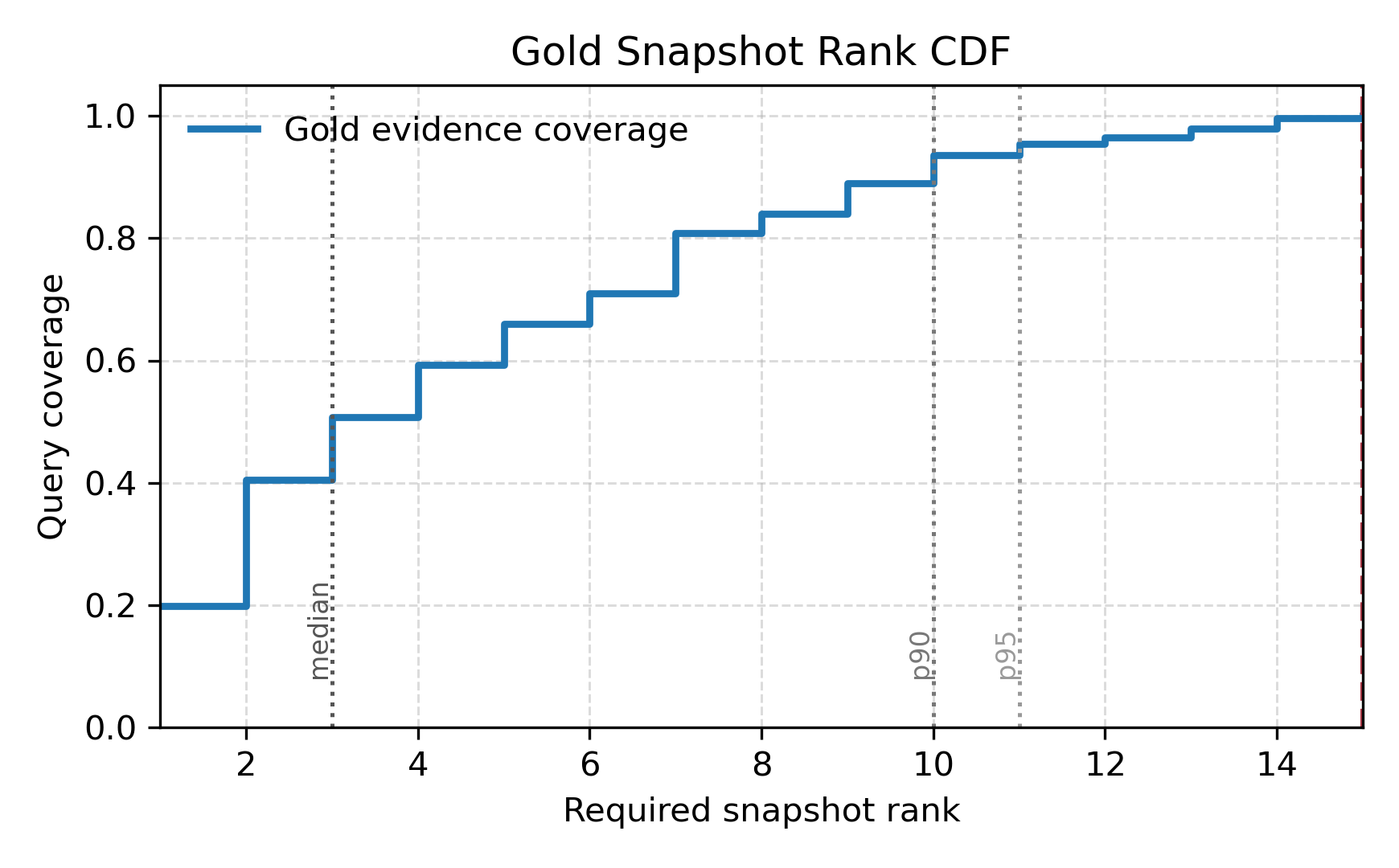}
\end{minipage}\hfill
\begin{minipage}[t]{0.495\linewidth}
  \centering
  \includegraphics[width=\linewidth]{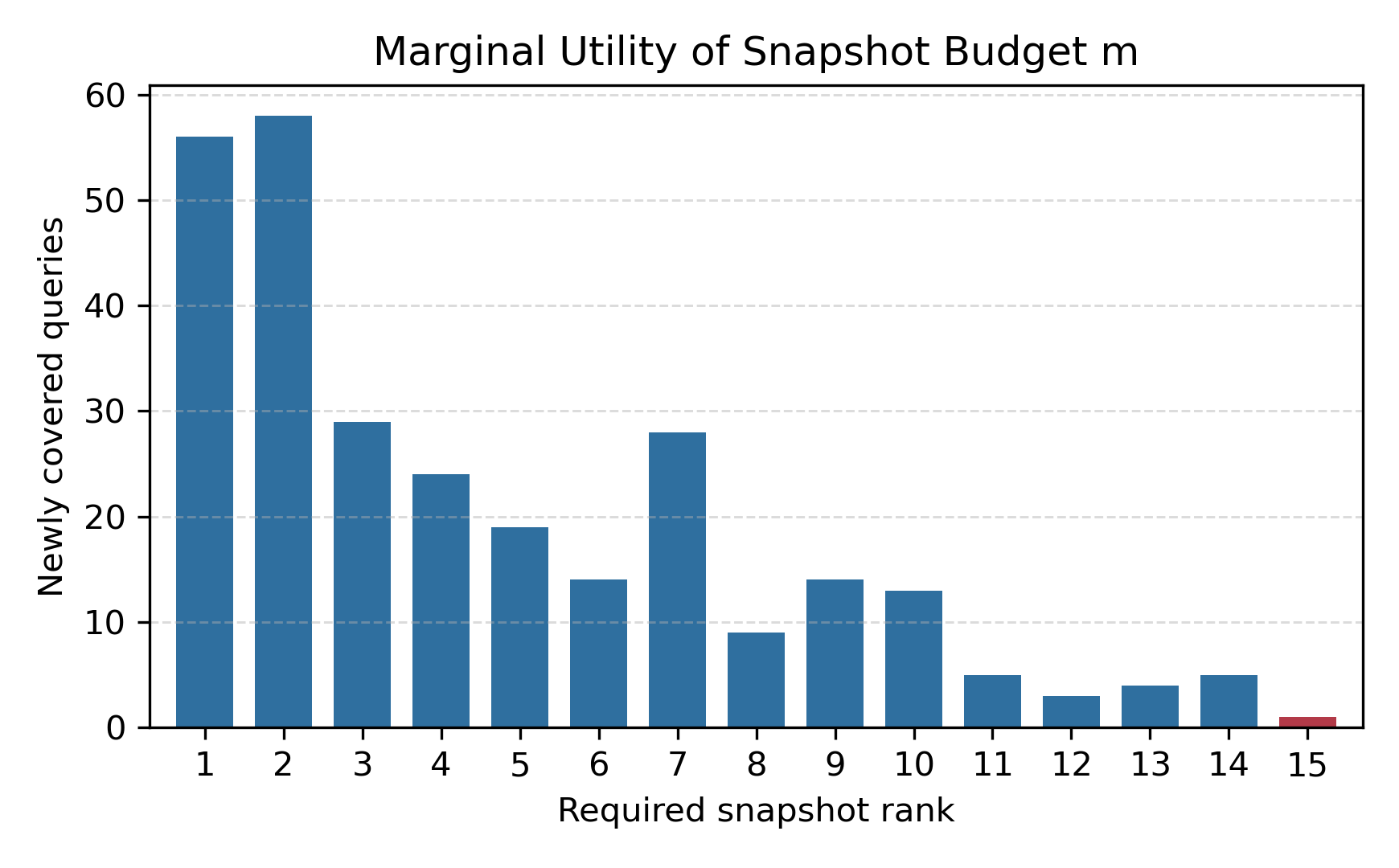}
\end{minipage}
\captionof{figure}{
\textbf{Snapshot Budget Analysis} for trajectory length \(m\) on the LoCoMo. The left plot shows the cumulative fraction of queries whose gold evidence is covered within a given required snapshot rank. 
The right plot shows the marginal number of newly covered queries at each rank, indicating that most gains occur at shallow depths while \(m=15\) covers the remaining long-tail evidence.
}
\label{fig:snapshot_budget_m}
\par}
\vspace{0.5em}
}

\newcommand{\EmpiricalResultsPageNineFloat}{%
\noindent\begin{minipage}{\linewidth}
\centering
\captionsetup{labelfont=bf,hypcap=false}
\begin{minipage}[t]{0.495\linewidth}
  \centering
  \includegraphics[width=\linewidth]{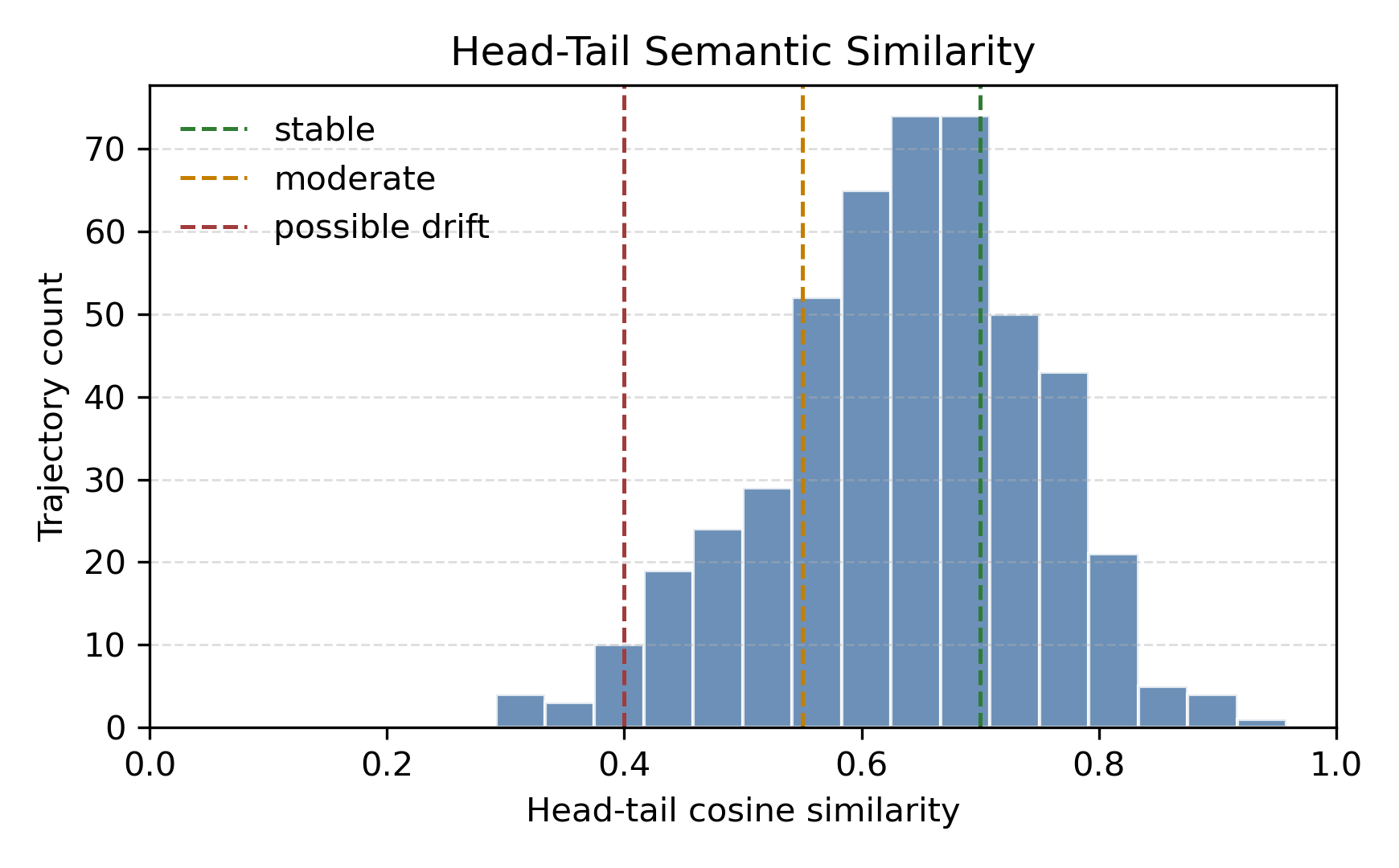}
\end{minipage}\hfill
\begin{minipage}[t]{0.495\linewidth}
  \centering
  \includegraphics[width=\linewidth]{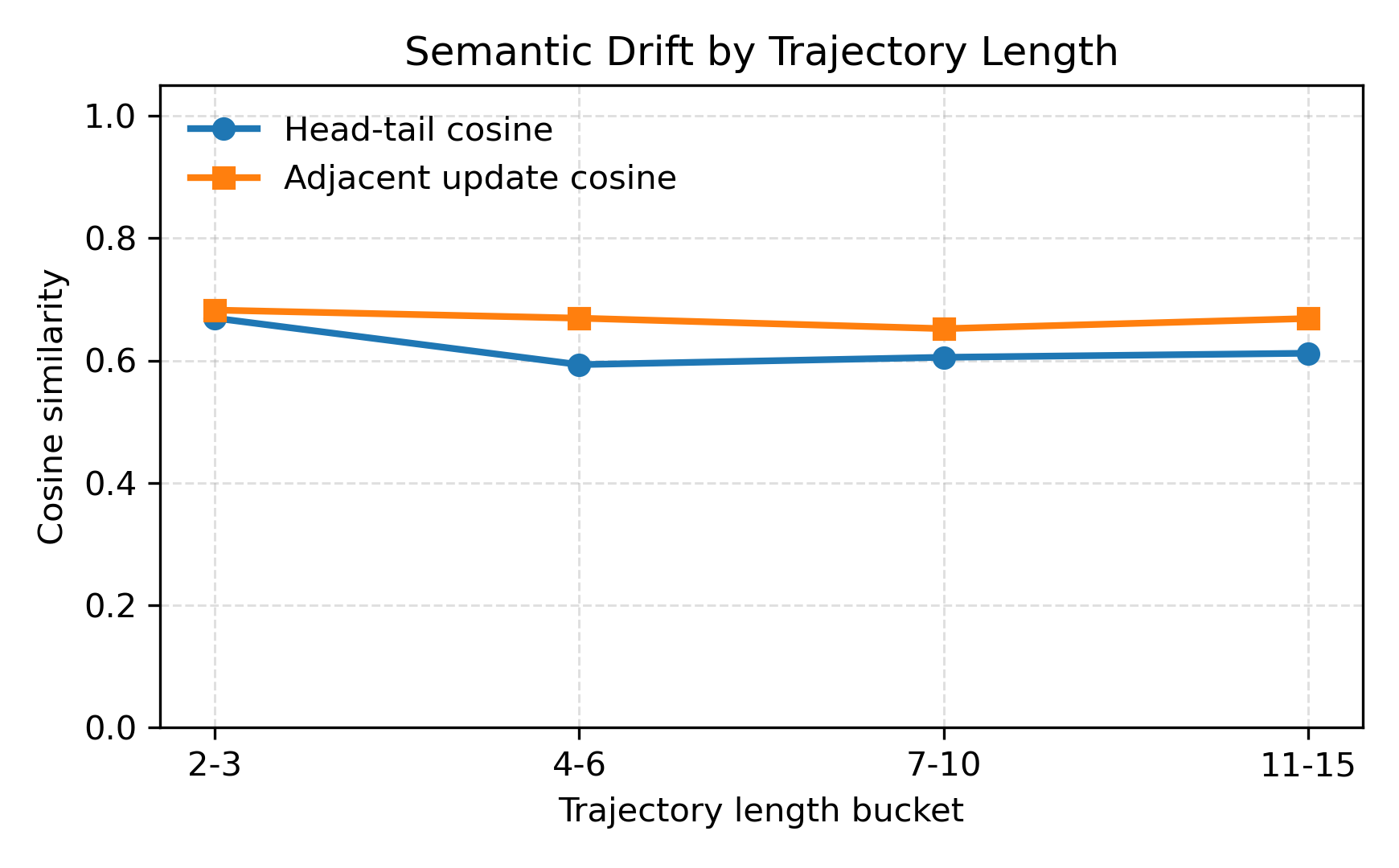}
\end{minipage}

\vspace{0.3em}
\captionof{figure}{\textbf{Trajectory Analysis.} The left panel shows head-tail semantic similarity for non-singleton trajectories, while the right panel breaks semantic drift down by trajectory length and contrasts global head-tail span with local adjacent-snapshot similarity. Most trajectories remain semantically coherent, and longer trajectories mainly increase global span rather than causing unstable local updates.}
\label{fig:trajectory_semantic_drift_analysis}
\end{minipage}
\vspace{0.3em}
}

\title{TrajWiki: Source-Grounded Memory Trajectories for Long-Horizon Dialogue Agents}

\author{%
Jingyu Sun\textsuperscript{1,2} \quad
Yuyang Xue\textsuperscript{3} \quad
Mingyang Li\textsuperscript{1} \quad
Zhengtao Yao\textsuperscript{4} \\
Jiachen Li\textsuperscript{5} \quad
Yang Cui\textsuperscript{1} \quad
Wenhao Cai\textsuperscript{1} \quad
Haozhe Liu\textsuperscript{1} \\
Fangying Wang\textsuperscript{1} \quad
Magdalene Katharina Montgomery\textsuperscript{2} \\
Syed Murtuza Baker\textsuperscript{1} \quad
Hongpeng Zhou\textsuperscript{1,*} \\[0.5em]
\normalfont\small\textsuperscript{1}The University of Manchester \quad
\textsuperscript{2}The University of Melbourne \\
\normalfont\small\textsuperscript{3}The University of Edinburgh \quad
\textsuperscript{4}University of Southern California \\
\normalfont\small\textsuperscript{5}The University of Texas at Austin \\
\normalfont\small\textsuperscript{*}Corresponding author
}

\hypersetup{%
  hidelinks,
  pdftitle={TrajWiki: Source-Grounded Memory Trajectories for Long-Horizon Dialogue Agents},
  pdfauthor={Jingyu Sun, Yuyang Xue, Mingyang Li, Zhengtao Yao, Jiachen Li, Yang Cui, Wenhao Cai, Haozhe Liu, Fangying Wang, Magdalene Katharina Montgomery, Syed Murtuza Baker, Hongpeng Zhou}
}

\begin{document}
\raggedbottom

\maketitle

\begin{abstract}
  Large language model agents have shown strong capabilities in generating coherent and contextually appropriate responses, yet robust long-horizon dialogue remains limited by the lack of external memory that is traceable, updatable, and diagnostically transparent. Existing memory-augmented agents often store memories as isolated records or overwritable states, making it difficult to preserve how information originates, evolves, conflicts, or becomes obsolete over time. We propose \textbf{TrajWiki}, a trajectory-based memory framework for long-horizon conversational agents. Instead of treating memory as static entries, TrajWiki represents each memory as a source-grounded evolution trajectory, maintained through immutable episodic snapshots and claim-level operations such as \texttt{ADD}, \texttt{REVISE}, and \texttt{DEPRECATE}. To reduce fragmentation and retrieval cost, TrajWiki further introduces \textbf{Memory Wiki}, a persistent intermediate layer that incrementally compiles dialogue history into structured and interlinked wiki pages capturing salient entities, events, quantities, topics, and conflicts. At inference time, queries are routed hierarchically from relevant wiki pages to linked memory trajectories, then to corresponding snapshots and source messages for evidence-grounded answer synthesis. Experiments on LoCoMo and MedMT show that TrajWiki improves long-horizon dialogue performance across both open-source and closed-source LLM backbones, while providing greater interpretability and diagnostic visibility into memory evolution, retrieval failures, and answer generation.Code is available at \url{https://github.com/JingyuSunUOE/TrajWiki}.
\end{abstract}

\section{Introduction}

\afterpage{%
\begin{figure}[!t]
  \centering
  \captionsetup{labelfont=bf}
  \includegraphics[width=0.80\linewidth]{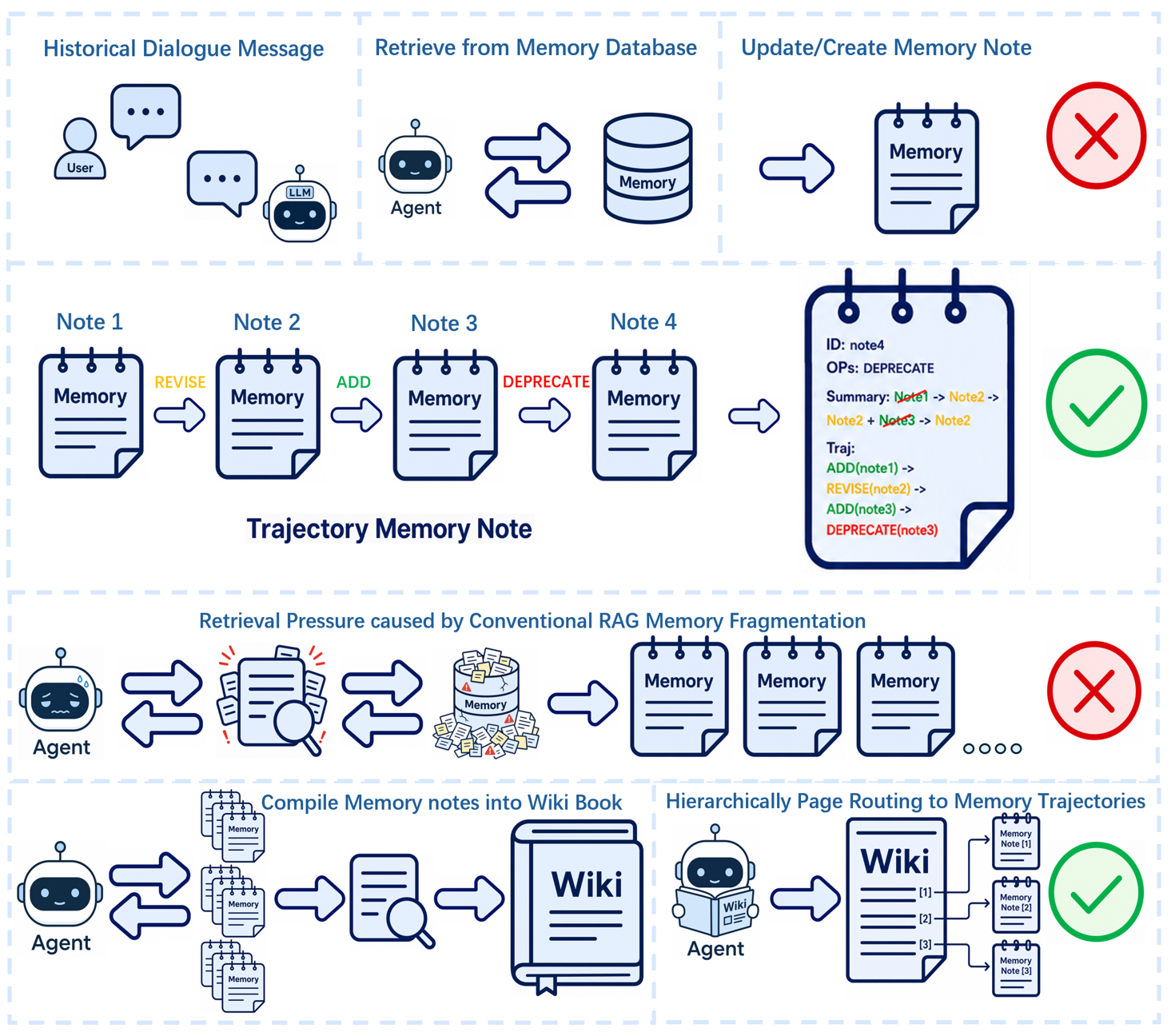}
  \caption{\textbf{Comparison of conventional agent memory and TrajWiki's source-grounded trajectory memory and Memory Wiki.} Conventional memory retains only the current belief and retrieves flat, isolated memories. TrajWiki preserves source-grounded update trajectories and structures them into a Memory Wiki, enabling reasoning over how facts evolve, conflict, and relate across long-horizon dialogue.}\label{fig:content}
\end{figure}
}

Large language model (LLM) agents are systems built around large language models that autonomously plan, reason, and act by iteratively interacting with users, external tools, and environments to solve complex, multi-step tasks, and they have shown early success in domains such as coding, research assistance, and broader task automation \citep{chowa2026language, yao2022react, schick2023toolformer, wang2023voyager, yuan2025poact, wan2025compass, rawat2025pre, gutierrez2024hipporag}. These systems rely on maintaining and updating context across successive interaction steps to support coherent decision-making; however, sustaining coherent long-horizon interaction remains difficult because the underlying models are constrained by finite context windows \citep{dai2019transformer, lidayan2025abbel, shahnovsky2026ai}. This limitation becomes particularly acute in long-term dialogue, where useful responses often depend on user-specific facts, evolving preferences, prior commitments, temporally distant events, and evidence scattered across multiple sessions \citep{zhao2025llms, joko2024doing, zhang2025towards, guo2026towards, maharana2024evaluating, wu2024longmemeval}. Simply extending the prompt with more history is costly and brittle, and does not provide reliable mechanisms for updating, organizing, or diagnosing long-term memory \cite{wang2023augmenting, zhong2024memorybank, kang2025memory, chhikara2025mem0, xu2025mem}.

Recent work addresses this limitation by equipping LLM agents with explicit external memory. Most memory-augmented systems follow a write/read paradigm: dialogue history is compressed into memory units such as notes, facts, user profiles, or event records, and relevant units are later retrieved to support generation \citep{park2023generative}. Other approaches improve memory organization through adaptive structures, dynamic linking, or graph-based retrieval \citep{xu2025mem}. While these methods demonstrate the value of persistent memory, we argue that two challenges remain underexplored, as shown in Figure~\ref{fig:content}. First, existing systems often retain only the current memory state without explicitly preserving its source-grounded update history. A retrieved memory may state what the agent currently believes, but not how that belief was formed, which claims were revised or deprecated, or which source utterances justify each update. Second, as fine-grained memories accumulate, flat retrieval over isolated memory items becomes increasingly fragmented, especially for temporal, multi-hop, or update-sensitive questions. Conventional retrieval-augmented generation treats knowledge mainly as searchable passages \citep{lewis2020retrieval}, whereas long-horizon dialogue memory also requires modeling how facts evolve, conflict, and relate over time.

In this paper, we propose \textbf{TrajWiki}, an interlinked, wiki-style, trajectory-based memory framework for long-horizon conversational agents. TrajWiki addresses the two challenges above through two key mechanisms. First, rather than representing memory as isolated and static records, it models each memory as a source-grounded evolutionary trajectory in which newly generated dialogue segments are stored as immutable episodic snapshots and linked to prior memory states through claim-level operations. Second, instead of repeatedly retrieving relevant information from fragmented raw chunks at query time, it reduces memory fragmentation by organizing trajectories into structured and interlinked wiki pages centered on entities, topics, events, and discrete factual units. At inference time, TrajWiki performs hierarchical retrieval: a query is first routed to relevant wiki pages, then to associated memory trajectories, and finally to episodic snapshots and original source messages, enabling evidence-grounded response generation. In summary, our contributions are fourfold:
\begin{itemize}
  \item We propose a source-grounded memory trajectory model that records long-term memory evolution through immutable episodic snapshots and claim-level update operations, preserving both the current effective memory and the historical path through which it was produced and making memory evolution inspectable at the level of claims and supporting evidence.
  \item We introduce an intermediate knowledge organization layer, \textbf{Memory Wiki}, that compiles trajectories into structured and interlinked wiki pages for more diagnosable retrieval, while preserving links to trajectories and source messages.
  \item We further introduce a hierarchical retrieval mechanism over interlinked memory trajectories that separates semantic organization, temporal memory evolution, and fine-grained evidence verification, reducing repeated reasoning over raw dialogue history while retaining provenance for diagnosis.
  \item We evaluate TrajWiki on LoCoMo and MedMT-Bench \citep{maharana2024evaluating, yang2026medmt}, covering both open-domain very long-term conversation and domain-specific long multi-turn medical dialogue. Experiments with open-source and closed-source LLM backbones show that TrajWiki improves long-horizon dialogue performance over strong memory-augmented baselines and provides richer diagnostic visibility into memory evolution, retrieval failures, and answer generation.
\end{itemize}

\section{Related Work}

\subsection{Agent Memory Systems}

External memory has become a central mechanism for LLM-based agents that operate beyond a fixed context window. Existing systems have explored multiple ways to store, update, and retrieve long-term interaction history. MemoryBank maintains user-specific memories with time- and importance-aware reinforcement, while MemGPT and MemoryOS frame memory management as a systems problem with hierarchical storage and context control \citep{packer2023memgpt, zhong2024memorybank, kang2025memory}. More recent work introduces richer memory structures, including hierarchical memory trees, temporal knowledge graphs, dynamically linked memory notes, and scalable conversational memory consolidation \citep{chhikara2025mem0, rezazadeh2024isolated, rasmussen2025zep, xu2025mem}. In long-horizon dialogue, timeline-based memory, reflective memory management, and graph-structured associative retrieval further demonstrate the value of structured memory for personalization and cross-session response generation \citep{ong2025towards, tan2025prospect, zhang2025bridging}. Despite these advances, most prior systems focus primarily on what to store and how to retrieve it, rather than explicitly modeling how memory states evolve. Even when memories are updated or linked, the update process is rarely represented as a provenance-aware sequence of immutable snapshots with claim-level edit semantics. This makes it difficult to inspect which claims were added, revised, or deprecated, and which source interactions justified each change. 

\subsection{Retrieval-Augmented Generation}

Retrieval-augmented generation (RAG) grounds language generation in external non-parametric memory by retrieving relevant evidence and conditioning generation on the retrieved content \citep{lewis2020retrieval}. While standard RAG often retrieves flat top-$k$ passages from relatively static corpora, recent methods improve retrieval with additional structure and multi-stage reasoning. Iter-RetGen interleaves retrieval and generation to refine evidence iteratively; RAPTOR organizes documents into a tree of recursive summaries; and GraphRAG builds graph-based indices to support global reasoning and query-focused summarization over large corpora \citep{shao2023enhancing, sarthi2024raptor, edge2024local}. TrajWiki shares the broader insight that retrieval benefits from structure, but targets a different setting. Existing RAG systems are mainly designed to query external document collections, whereas long-horizon dialogue agents must maintain an internal memory that evolves through ongoing interaction. In this setting, relevant evidence depends not only on semantic similarity, but also on when a memory was created, how it was revised, which earlier claims were superseded, and which source messages support the current state. 

\section{Methodology}

\afterpage{%
\begin{figure}[!t]
  \centering
  \captionsetup{labelfont=bf}
  \includegraphics[width=1\linewidth]{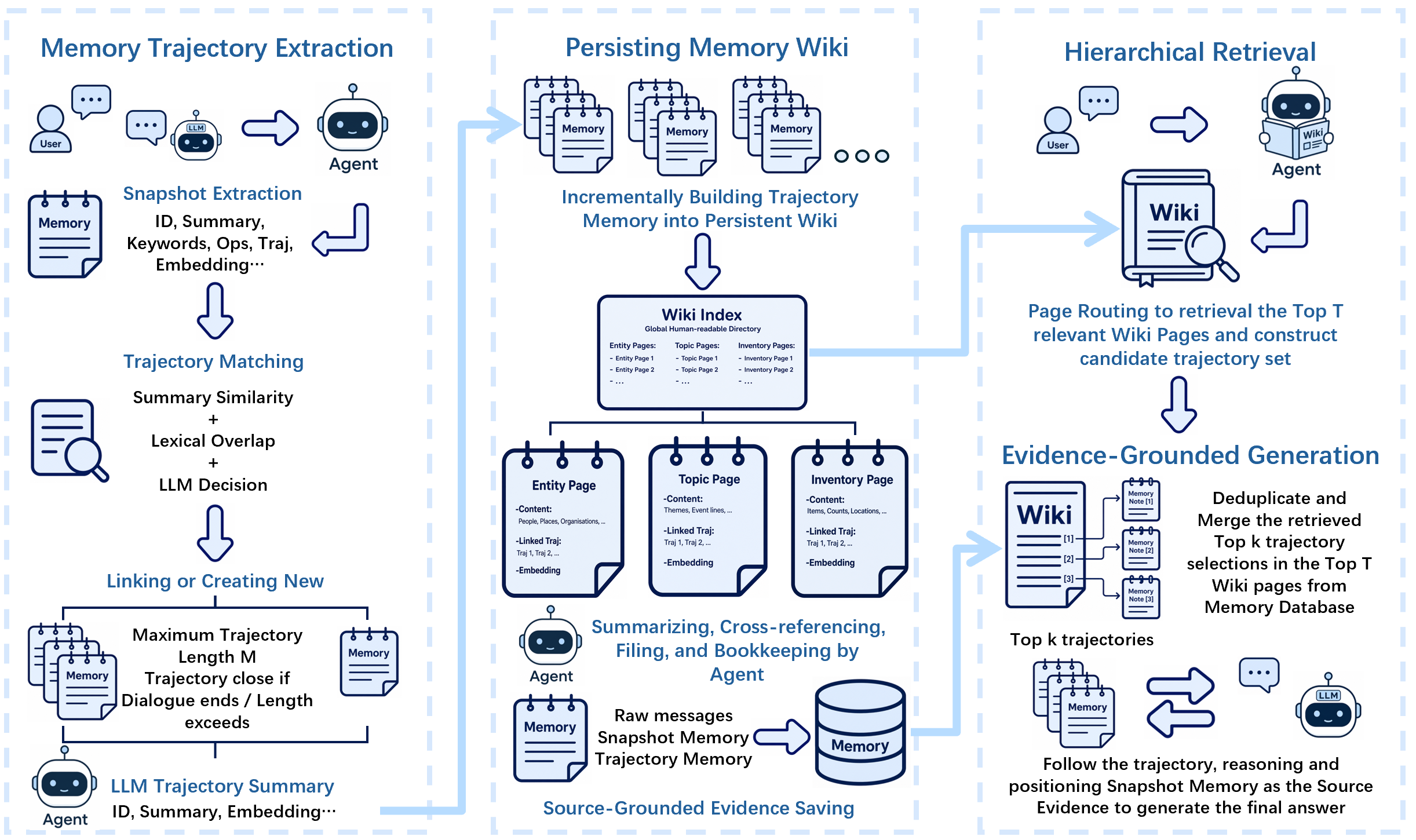}
  \caption{\textbf{Overview of the TrajWiki architecture.} TrajWiki converts dialogue history into source-grounded memory trajectories, compiles them into a persistent and interlinked Memory Wiki, retrieves evidence hierarchically from wiki pages to trajectories and source messages, and generates answers grounded in the retrieved evidence. The Memory Wiki serves as a compounding knowledge artifact, maintaining cross-references, flagged conflicts, and synthesized evidence as memory accumulates over time.}\label{fig:overview}
\end{figure}
}
\vspace{0.5em}

We propose \textbf{TrajWiki}, as shown in Figure~\ref{fig:overview}, a long-term memory framework for dialogue agents that represents memory as source-grounded trajectories and organizes them into a persistent Memory Wiki. Unlike standard retrieval-augmented generation, which retrieves external passages at inference time \citep{lewis2020retrieval}, TrajWiki first converts dialogue history into evolving memory trajectories, then compiles them into wiki-style pages for hierarchical retrieval and evidence-grounded answer generation.

Given a dialogue history
\begin{equation}
\mathcal{D}=\{m_1,\ldots,m_N\}, \qquad m_i=(x_i,u_i,\tau_i,\rho_i),
\end{equation}
where \(x_i\), \(u_i\), \(\tau_i\), and \(\rho_i\) denote message content, speaker, timestamp, and source reference, respectively, TrajWiki follows the pipeline
\begin{equation}
\mathcal{D} \rightarrow \Sigma \rightarrow \mathcal{T} \rightarrow \mathcal{W} \rightarrow \mathcal{E}_q \rightarrow a_q .
\end{equation}
Here, \(\Sigma\) denotes episodic snapshots, \(\mathcal{T}\) memory trajectories, \(\mathcal{W}\) the Memory Wiki, \(\mathcal{E}_q\) the retrieved evidence for query \(q\), and \(a_q\) the final answer. This design separates provenance tracking, persistent knowledge organization, and query-time grounding.

\subsection{Provenance-Aware Memory Trajectories}

TrajWiki stores memory as source-linked trajectories rather than isolated records. Each dialogue exchange is converted into an immutable episodic snapshot containing extracted claims, speaker and temporal context, and links to raw source messages. A memory trajectory is an ordered append-only sequence:
\begin{equation}
T_j = (\sigma_{j,1}, \sigma_{j,2}, \ldots, \sigma_{j,n_j}),
\end{equation}
where each snapshot \(\sigma_{j,t}\) remains traceable to the messages that produced it.

For a new snapshot \(\sigma\), TrajWiki either appends it to the most compatible trajectory or initializes a new one:
\begin{equation}
T^\star=\arg\max_{T_j\in\mathcal{T}} g(\sigma,T_j),
\qquad
\sigma \mapsto
\begin{cases}
T^\star, & g(\sigma,T^\star)\geq \delta,\\
T_{\mathrm{new}}, & \text{otherwise}.
\end{cases}
\end{equation}
The compatibility function \(g\) considers entity continuity, facet overlap, lexical evidence, temporal compatibility, and semantic drift, allowing related updates to be connected without collapsing unrelated facts into overly broad trajectories.

Within each trajectory, memory evolution is represented through claim-level edit operations:
\begin{equation}
C_{j,t}=
\operatorname{Apply}(C_{j,t-1},\mathcal{O}_{j,t}),
\qquad
\mathcal{O}_{j,t}\subseteq
\{\texttt{ADD},\texttt{REVISE},\texttt{DEPRECATE}\}.
\end{equation}
Thus, revised or deprecated claims are not deleted; they remain linked to their original snapshots and source references, making memory updates inspectable over time.

\subsection{Persistent Memory Wiki}

Although trajectories preserve provenance, directly retrieving over all snapshots and claims can become fragmented as memory grows. TrajWiki therefore compiles trajectories into a persistent \textbf{Memory Wiki}, which serves as a readable and routable organization layer between raw memory and query-time retrieval.

Each wiki page is represented as
\begin{equation}
p=(\eta,y,h,L_p), \qquad L_p\subseteq\mathcal{T},
\end{equation}
where \(\eta\) is the page type, \(y\) is the title, \(h\) is the page content, and \(L_p\) is the set of linked trajectories. We use four page types:
\[
\eta\in\{\texttt{index},\texttt{entity},\texttt{topic},\texttt{inventory}\}.
\]
The \texttt{index} page provides a global directory, \texttt{entity} pages organize recurring entities, \texttt{topic} pages capture events or thematic clusters, and \texttt{inventory} pages collect list-like or countable facts.

Wiki construction is summarized as
\begin{equation}
\mathcal{W}=
\operatorname{Compile}
(\mathcal{T}; C^{\mathrm{active}}, C^{\mathrm{hist}}, A),
\end{equation}
where \(C^{\mathrm{active}}\) and \(C^{\mathrm{hist}}\) denote active and historical claims, and \(A\) denotes source anchors and metadata. The wiki does not replace source evidence with summaries; instead, it organizes accumulated memory while preserving links to trajectories and raw messages. To avoid making trajectories reachable only through the global directory, TrajWiki encourages non-index coverage:
\begin{equation}
\bigcup_{p\in\mathcal{W}_{\mathrm{non\text{-}index}}} L_p \approx \mathcal{T}.
\end{equation}

\subsection{Hierarchical Retrieval and Evidence-Grounded Generation}

Given a query \(q\), TrajWiki retrieves evidence through a hierarchy of wiki pages, trajectories, and source-linked snapshots. It first routes the query to relevant wiki pages and uses their links to construct a candidate trajectory set:
\begin{equation}
\mathcal{P}_q=
\operatorname{Top}_t\{S_p(q,p):p\in\mathcal{W}\},
\qquad
\mathcal{C}_q=\bigcup_{p\in\mathcal{P}_q}L_p .
\end{equation}
Here, \(S_p\) combines semantic, lexical, entity, facet, and temporal signals. This stage narrows retrieval from the entire memory store to organized memory regions.

From \(\mathcal{C}_q\), TrajWiki selects trajectories using a relevance-and-coverage objective:
\begin{equation}
\mathcal{R}_q=
\operatorname{Top}_k
\left\{
b(q,T)+\alpha\Delta_{\mathrm{cov}}(T)-\lambda\operatorname{red}(T):
T\in\mathcal{C}_q
\right\}.
\end{equation}
The base score \(b(q,T)\) measures relevance, \(\Delta_{\mathrm{cov}}\) rewards complementary entities, events, facets, or countable items, and \(\operatorname{red}\) penalizes redundancy. This encourages retrieval of multiple complementary trajectories for list, count, temporal, and multi-hop questions.

Selected trajectories are expanded into provenance-linked evidence:
\begin{equation}
\mathcal{E}_q=
\{(T,\sigma,c,\rho,\tau):
T\in\mathcal{R}_q,\,
\sigma\in T,\,
c\in C_T\}.
\end{equation}
Thus, retrieved evidence contains not only text fragments, but also their trajectory, claim, source reference, and timestamp.

The final answer is generated under a source-support constraint:
\begin{equation}
z_q=\operatorname{Synth}_{\theta}(q,\mathcal{E}_q),
\qquad
a_q=\operatorname{Answer}(z_q),
\qquad
\operatorname{refs}(a_q)\subseteq\operatorname{refs}(\mathcal{E}_q).
\end{equation}
The synthesis \(z_q\) records the answerability decision, supporting facts, source references, temporal anchors, and uncertainties. If the retrieved evidence is insufficient, TrajWiki abstains or performs a controlled retry by generating retrieval hints and rerouting through the wiki and trajectory layers. In all cases, the final answer must remain grounded in retrieved source evidence.

Overall, TrajWiki decomposes long-term dialogue memory into three layers: source-grounded trajectories for memory evolution, Memory Wiki pages for persistent organization, and hierarchical retrieval for query-time grounding. This enables agents to accumulate, revise, and query long-term memory while preserving the provenance needed to audit both memory updates and generated answers. We provide the reproducible implementation details, including deterministic scoring features, structured-output schemas, hyperparameters, validation rules, and retry/repair procedures, in Appendix~\ref{app:reproducible_details}.

\afterpage{%
\begin{table}[!t]
\centering
\captionsetup{labelfont=bf}
\caption{\textbf{Overall performance comparison on LoCoMo and MedMT-Bench.} 
For LoCoMo, we report F1, BLEU, and Acc separately for each question type. For MedMT-Bench, we report Acc for each subset. The best performance is marked in bold. }
\label{tab:overall_results}
\scriptsize
\setlength{\tabcolsep}{2.2pt}
\renewcommand{\arraystretch}{1.08}
\resizebox{\textwidth}{!}{%
\begin{tabular}{l l|ccc|ccc|ccc|ccc|c|c|c}
\hline
\multirow{3}{*}{\textbf{Model}} &
\multirow{3}{*}{\textbf{Method}} &
\multicolumn{12}{c|}{\textbf{LoCoMo}} &
\multicolumn{3}{c}{\textbf{MedMT-Bench}} \\
\cline{3-17}
& &
\multicolumn{3}{c|}{\shortstack{\textbf{Multi}\\\textbf{Hop}}} &
\multicolumn{3}{c|}{\textbf{Temporal}} &
\multicolumn{3}{c|}{\shortstack{\textbf{Open}\\\textbf{Domain}}} &
\multicolumn{3}{c|}{\shortstack{\textbf{Single}\\\textbf{Hop}}} &
\textbf{LCMU} &
\textbf{RCI} &
\textbf{IC} \\
\cline{3-17}
& &
\textbf{F1} & \textbf{BLEU} & \textbf{Acc} &
\textbf{F1} & \textbf{BLEU} & \textbf{Acc} &
\textbf{F1} & \textbf{BLEU} & \textbf{Acc} &
\textbf{F1} & \textbf{BLEU} & \textbf{Acc} &
\textbf{Acc} &
\textbf{Acc} &
\textbf{Acc} \\
\hline

\multirow{6}{*}{\ModelCell{GPT}{4o-mini}}
& Full Context
& 28.11 & 21.71 & 32.98 & 23.11 & 16.12 & 30.22 & 16.76 & 13.57 & 35.42 & 43.77 & 36.43 & 71.72
& 27.12 & 11.16 & 7.47 \\
& Naive RAG
& 21.78 & 17.08 & 20.21 & 20.35 & 15.37 & 24.92 & 14.89 & 11.67 & 26.04 & 40.49 & 34.13 & 60.52
& 16.43 & 18.86 & 7.82 \\
& LangMem
& 22.01 & 16.92 & 28.01 & 20.59 & 14.43 & 31.78 & 10.31 & 8.09 & 23.96 & 34.4 & 26.62 & 55.28
& 17.78 & 16.34 & 8.26 \\
& A-MEM
& 27.02 & 20.09 & 37.58 & 41.52 & 35.6 & 26.17 & 12.44 & 9.97 & 23.96 & 38.74 & 34.03 & 50.54
& 12.51 & 17.12 & 9.33 \\
& Mem0
& 34.52 & 25.13 & 40.07 & \textbf{44.73} & 37.86 & 30.56 & 13.12 & 9.5 & 25.02 & 37.19 & 33.64 & 44.48
& 17.11 & 14.52 & 12.17 \\
\rowcolor{gray!18}
& \textbf{TrajWiki}
& \textbf{35.23} & \textbf{36.47} & \textbf{59.44} & 42.42 & \textbf{41.14} & \textbf{35.39} & \textbf{20.56} & \textbf{18.66} & \textbf{42.12} & \textbf{48.72} & \textbf{43.14} & \textbf{78.56}
& \textbf{29.33} & \textbf{21.42} & \textbf{17.68} \\
\hline

\multirow{6}{*}{\ModelCell{Qwen3}{32B}}
& Full Context
& 27.79 & 20.91 & 49.65 & 32.47 & 26.56 & 32.27 & 15.52 & 11.44 & 39.67 & 29.53 & 32.71 & \textbf{79.68}
& 29.63 & 13.16 & 9.42 \\
& Naive RAG
& 24.22 & 19.01 & 35.82 & 22.61 & 18.21 & 20.25 & 12.88 & 8.7 & 36.46 & 35.19 & 27.27 & 69.08
& 19.63 & 21.05 & 7.62 \\
& LangMem
& 26.01 & 24.37 & 38.16 & 24.9 & 23.81 & 20.28 & 9.15 & 6.87 & 22.92 & 36.61 & 34.88 & 44.15
& 18.18 & 16.67 & 7.74 \\
& A-MEM
& 25.17 & 19.12 & 41.48 & 29.79 & 24.63 & 28.78 & 11.29 & 7.68 & 41.67 & 33.76 & 26.21 & 63.26
& 13.89 & 17.07 & 10.36 \\
& Mem0
& 26.32 & 20.17 & 45.74 & 32.41 & 29.68 & 31.49 & 15.03 & 11.28 & \textbf{42.7} & \textbf{42.58} & 35.15 & 73.72
& 17.89 & 23.68 & 11.21 \\
\rowcolor{gray!18}
& \textbf{TrajWiki}
& \textbf{31.42} & \textbf{30.56} & \textbf{50.33} & \textbf{33.48} & \textbf{31.26} & \textbf{37.33} & \textbf{17.86} & \textbf{16.44} & 39.68 & 41.88 & \textbf{40.42} & 78.34
& \textbf{33.42} & \textbf{27.33} & \textbf{15.34} \\
\hline

\multirow{6}{*}{\ModelCell{Qwen3}{8B}}
& Full Context
& \textbf{39.05} & 31.16 & 39.36 & 33.64 & 27.19 & 26.17 & \textbf{20.67} & \textbf{16.52} & 31.54 & 43.27 & \textbf{43.45} & \textbf{78.21}
& 26.69 & 10.22 & 6.76 \\
& Naive RAG
& 27.25 & 21.05 & 29.79 & 27.86 & 22.96 & 18.69 & 12.18 & 8.98 & 38.54 & \textbf{46.32} & 41.44 & 64.8
& 16.32 & 18.32 & 6.42 \\
& LangMem
& 16.48 & 15.08 & 27.09 & 17.85 & 15.83 & 15.61 & 9.47 & 6.96 & 24.58 & 36.24 & 35.16 & 51.77
& 17.81 & 15.47 & 7.74 \\
& A-MEM
& 27.54 & 22.08 & 35.46 & \textbf{34.30} & 28.38 & 21.18 & 11.39 & 8.62 & 22.29 & 42.3 & 37.66 & 59.45
& 12.04 & 16.32 & 8.76 \\
& Mem0
& 31.73 & 24.82 & 38.27 & 28.96 & 26.24 & 20.16 & 15.03 & 11.28 & 21.18 & 42.58 & 35.15 & 57.64
& 15.42 & 13.62 & 10.52 \\
\rowcolor{gray!18}
& \textbf{TrajWiki}
& 30.25 & \textbf{32.14} & \textbf{42.14} & 31.48 & \textbf{30.48} & \textbf{34.62} & 16.89 & 13.52 & \textbf{39.78} & 43.46 & 37.24 & 63.48
& \textbf{27.56} & \textbf{20.11} & \textbf{14.28} \\
\hline

\end{tabular}%
}
\end{table}
}

\section{Experiment}

\subsection{Datasets and Evaluation}

We evaluate \textbf{TrajWiki} on two long-horizon dialogue benchmarks.
Our primary benchmark is LoCoMo, a long-term conversational memory
dataset with gold question-answer pairs
\citep{maharana2024evaluating}. We use its QA task, which covers
\textit{single-hop}, \textit{multi-hop} requiring cross-session
evidence aggregation, \textit{temporal reasoning}, and \textit{open-domain /
world-knowledge} questions. Following Mem0
\citep{chhikara2025mem0}, we exclude the adversarial category to ensure
fair comparison with prior memory-augmented baselines. To assess
robustness beyond open-domain conversations, we further use MedMT-Bench
as a domain-specific stress test \citep{yang2026medmt}. MedMT-Bench
evaluates long multi-turn medical conversations using instance-level
rubrics rather than a single reference answer.
Since our goal is to evaluate memory mechanisms rather than medical
knowledge or multimodal reasoning, we select memory-relevant cases from
three categories: \textit{Long-Context Memory and Understanding (LCMU)},
\textit{Resistance to Contextual Interference (RCI)}, and the
\textit{Information Contradiction (IC)} subset of
\textit{Instruction Clarification}. These categories test long-history
alignment, robustness to contextual noise, and conflict detection
between historical facts and current user statements. 

We compare TrajWiki with five baselines: \textbf{Full Context}, which
directly conditions on the complete dialogue history; \textbf{Naive
RAG}, evaluated with both session-level and turn-level retrieval
granularity \citep{lewis2020retrieval}; \textbf{LangMem}, which
extracts salient conversational information for persistent memory
retrieval \citep{langmem2025}; \textbf{A-MEM}, which dynamically links
structured memory notes \citep{xu2025mem}; and \textbf{Mem0}, which
performs incremental memory extraction, consolidation, and retrieval
\citep{chhikara2025mem0}. For LoCoMo, we report F1 and
BLEU-1 as lexical overlap metrics \citep{papineni2002bleu}. Since
long-horizon dialogue answers may be semantically correct despite
different surface forms, we additionally use a strict LLM-as-a-Judge protocol
to evaluate answer correctness \citep{zheng2023judging}. For
MedMT-Bench, we follow its rubric-based setting and report test-point
pass accuracy. Details of metric computation are
provided in Appendix~\ref{app:metrics}.

\afterpage{\EmpiricalResultsPageEightFloats}
\afterpage{\afterpage{\EmpiricalResultsPageEightTable}}
\afterpage{\afterpage{\afterpage{\EmpiricalResultsPageNineFloat}}}

\subsection{Implementation Details}

For all baselines, we follow their original configurations, hyperparameters, and system prompts. When details are underspecified, we use the closest publicly described setting while keeping the backbone model, answering format, and evaluation protocol consistent across methods. For closed-source experiments, we use \texttt{GPT\mbox{-}4o\mbox{-}mini} as the backbone model and LLM judge. For open-source experiments, we use \texttt{Qwen3\mbox{-}8B} and \texttt{Qwen3\mbox{-}32B} in non-thinking mode as backbone models, and \texttt{Qwen3\mbox{-}8B} as the judge. TrajWiki uses \texttt{Qwen3\mbox{-}Embedding\mbox{-}8B} for all embedding-based retrieval components. We report all major prompts in Appendix~\ref{app:prompts}. For TrajWiki, the default setting uses maximum trajectory length \(m=15\), routes each query through the top \(t=15\) wiki pages, and selects the top \(k=15\) trajectories. These settings balance retrieval coverage and computational cost.

\subsection{Empirical Results}

Table~\ref{tab:overall_results} reports the main results on LoCoMo and MedMT-Bench \citep{maharana2024evaluating, yang2026medmt}. Across both benchmarks and both GPT- and Qwen-based backbones, \textbf{TrajWiki} consistently outperforms most of the baselines, including Full Context, Naive RAG, LangMem, A-MEM, and Mem0 \citep{langmem2025, xu2025mem, chhikara2025mem0}. On LoCoMo, TrajWiki achieves better performance and improves over the strongest memory baseline, Mem0. The gains are most pronounced on multi-hop questions, where answering requires aggregating evidence distributed across sessions. This suggests that source-grounded trajectories and Memory Wiki routing help recover complementary memory traces beyond local semantic similarity. TrajWiki also shows a consistent advantage on temporal reasoning questions, although the margin is smaller, reflecting the difficulty of resolving temporally distant and evolving dialogue evidence. On MedMT-Bench, the performance gap becomes more substantial. Most baselines degrade in this domain-specific setting, especially on the Information Contradiction subset, where models must detect conflicts between historical facts and the current user statement. In contrast, TrajWiki remains robust across the three medical scenarios, demonstrating that its memory mechanism transfers beyond open-domain conversations to high-constraint professional dialogues. We attribute this improvement to two aspects of the design: trajectory-level memory preserves how claims evolve over time, while wiki-based routing reduces retrieval fragmentation before evidence-level expansion. 

\subsection{Hyperparameter Analysis}

We analyze three key hyperparameters of TrajWiki on LoCoMo: the maximum trajectory length \(m\), the number of routed wiki pages \(t\), and the number of selected trajectories \(k\). Rather than rerunning answer generation for every setting, we use the compact retrieval diagnostics saved during benchmarking to perform an offline coverage analysis. Figure~\ref{fig:retrieval_cutoffs} compares the effects of page routing cutoff and trajectory top-\(k\). Increasing \(t\) substantially improves the probability that gold trajectories enter the page-induced candidate universe: small values such as \(t=5\) or \(t=10\) miss many supporting trajectories, while \(t=15\) provides a practical trade-off and \(t=20\) further improves coverage at the cost of a larger candidate set. Increasing \(k\) similarly improves gold trajectory coverage, but also increases the amount of evidence passed to answer generation, which may introduce additional noise. Figure~\ref{fig:snapshot_budget_m} analyzes the required snapshot depth inside gold trajectories. Most queries require only shallow trajectory history, while a small long tail requires deeper snapshots. The marginal utility plot shows that most additional coverage is gained at small ranks, with diminishing returns after \(m=10\), while \(m=15\) covers the remaining long-tail evidence. These results support the design of TrajWiki as a hierarchical memory system: wiki routing controls retrieval breadth, trajectory selection controls evidence depth, and bounded trajectories preserve sufficient historical context without requiring unbounded memory expansion.

\subsection{Counterfactual Ablation}

We conduct a counterfactual ablation on the LoCoMo multi-hop split using GPT-4o-mini. The goal is to isolate how TrajWiki's memory organization affects evidence retrieval, grounding, and context efficiency. Since the counterfactual variants are derived from a single full TrajWiki run and do not rerun answer generation, we report retrieval and context-grounding metrics for the ablated variants; observed answer-level metrics are available only for the full TrajWiki run. A more detailed budget-controlled analysis is provided in Appendix~\ref{app:offline_ablation}. Table~\ref{tab:offline_ablation_main} shows the main results at \(k=15\) under a 32K offline context budget. Removing the Memory Wiki and directly ranking trajectories increases the candidate universe from 55.6 to 130.4 trajectories, while reducing gold source-reference coverage from 0.610 to 0.356. This suggests that wiki routing provides useful semantic organization rather than acting as a simple lossy filter. The wiki-only setting further shows that summaries alone are insufficient for source-grounded QA: although they can point to relevant trajectories, they do not recover gold source references. Flat raw-memory retrieval also has a much larger candidate universe and higher unsupported-evidence risk. Overall, the results support TrajWiki's combination of persistent wiki routing, source-grounded trajectories, and snapshot-level evidence expansion.

\subsection{Trajectory Semantic Drift Analysis}
We further analyze whether memory trajectories remain semantically coherent as they accumulate updates over time. For each trajectory, we compute the cosine similarity between the embeddings of its first and latest snapshots, which measures the semantic span of the trajectory, and the average cosine similarity between adjacent snapshots, which measures local continuity during memory updates. This analysis is performed offline using the saved snapshot embeddings from the LoCoMo run and does not involve additional model calls. As shown in Figure~\ref{fig:trajectory_semantic_drift_analysis}, the left panel indicates that most non-singleton trajectories maintain moderate to high head-tail similarity, while only a small fraction falls into the possible-drift region. The right panel further shows that longer trajectories naturally cover a broader semantic range, leading to lower head-tail similarity, but their adjacent-update similarity remains relatively stable. This suggests that TrajWiki trajectories support gradual memory evolution rather than arbitrary aggregation: even when a trajectory spans multiple related events, consecutive updates remain locally coherent. These results support the design of provenance-aware trajectories as a bounded yet extensible unit for long-term memory, where semantic drift can be measured and audited rather than hidden inside overwritten memory states.

\subsection{Computational Cost and Scalability Analysis}

We analyze the observed cost of TrajWiki on the LoCoMo using GPT-4o-mini. The cost is concentrated in memory construction and evidence organization rather than final answer generation: excluding benchmark-only evaluation, the run uses 43.28M deployment tokens, including 23.71M for reusable memory construction, 10.84M for query-time retrieval, and 6.92M for answer generation. This reflects TrajWiki's main trade-off: additional computation is used to build source-grounded trajectories and a persistent Memory Wiki. The scalability analysis further shows that the Memory Wiki reduces the query-time trajectory search space. Across dialogue samples, direct retrieval would consider 130.4 trajectories per query on average, while wiki routing reduces this to 55.6 candidate trajectories, a 2.35$\times$ reduction. Detailed phase-level costs, latency breakdowns, and memory-scaling results are provided in Appendix~\ref{app:cost_analysis}.

\subsection{Diagnostic Visibility and Limitations}
\label{sec:limitations}

TrajWiki preserves source links across snapshots, trajectories, Wiki pages, retrieved evidence, and final answers. We use \emph{diagnostic provenance} to denote observable evidence paths and associated risk indicators. These records support inspection of memory evolution and answer construction, but do not establish the correctness or completeness of either. The following analyses characterize the benefits of this visibility and the remaining limitations of the architecture.

\paragraph{Evidence retention and generation failures.}
TrajWiki depends on structured LLM outputs for extraction, memory updating, Wiki construction, answer synthesis, and validation. Malformed outputs can trigger expensive fallback and repair paths, particularly when models do not reliably follow structured-output instructions. Repeated summarization can also alter the salience of individual facts: retaining a source link does not ensure that the corresponding evidence remains prominent in the generated answer. In the original 282-query diagnostic run, the unsupported-answer-risk proxy was $0.610$, while mean gold-reference coverage in final answer support was only $0.332$. The former is an automated risk indicator, not a human-verified hallucination rate.

The diagnostic funnel on 200 queries further identifies where evidence is lost. Using each query's gold references as the denominator, mean retention was $100.0\%$ in raw memory, $81.7\%$ after Wiki reachability, $62.3\%$ after trajectory selection, approximately $62\%$ in the final source context, and $33.6\%$ in final answer support. Evidence is therefore lost both before retrieval is completed and when retrieved information is converted into answer support. These measurements describe source-reference retention, not independently verified factual correctness. Appendix~\ref{app:limitations} provides the underlying diagnostics and a case study of information loss through repeated summarization.

\paragraph{Exploratory human audit.}
We conducted a counterbalanced audit of 24 cases with two auditors who were not involved in system development. Each case was assessed under both the TrajWiki-packet and full-dialogue views, with different auditors assigned to the two views. Two developers adjudicated support and failure-stage labels before unblinding. Table~\ref{tab:limitations-human-audit} reports the results.

\begin{table}[t]
\centering
\small
\setlength{\tabcolsep}{4pt}
\caption{\textbf{Exploratory human audit of 24 cases.} Packet tokens and source rows are means; verification and localization scores are percentages. The reference support and failure-stage labels were adjudicated by two developers. K denotes thousands of tokens.}
\label{tab:limitations-human-audit}
\begin{tabular}{@{}lrrrrr@{}}
\toprule
Audit view & \shortstack{Packet\\tokens} & \shortstack{Source\\rows} & \shortstack{Support\\verification} & \shortstack{Exact\\stage} & \shortstack{Stage\\family} \\
\midrule
TrajWiki packet & 5.93K & 79.2 & 83.3 & 37.5 & 45.8 \\
Full dialogue   & 26.74K & 598.8 & 100.0 & 25.0 & 29.2 \\
\bottomrule
\end{tabular}
\end{table}

The full-dialogue view contained approximately $4.51\times$ as many tokens as the TrajWiki packet, while the packet yielded numerically higher exact-stage and stage-family localization scores. However, full dialogue provided better support verification, and exact-stage localization remained modest under both views. With only 24 cases and developer-adjudicated reference labels, this is preliminary evidence. The audit supports a more compact inspection surface with preliminary localization benefits.

\section{Conclusion}

We introduced \textbf{TrajWiki}, a long-term memory framework that represents dialogue history as provenance-aware memory trajectories and organizes them into a persistent Wiki-style semantic layer. By separating memory evolution and persistent organization, TrajWiki enables agents to accumulate, revise, and retrieve long-term conversational knowledge while preserving source-level evidence. Our experiments show that hierarchical retrieval over Wiki pages and trajectories improves the ability to locate multi-hop evidence, while evidence-grounded answer generation and post-hoc diagnostics make failures more interpretable. The results also highlight an important trade-off. TrajWiki incurs additional computational cost, but this cost supports reusable semantic organization, provenance tracking, and auditable retrieval, and can be amortized across future queries in long-term agent settings. Overall, TrajWiki demonstrates that treating memory as an evolving, source-grounded structure rather than a flat retrieval pool is a promising direction for building more reliable long-term dialogue agents.

\clearpage

\bibliographystyle{unsrtnat}
\bibliography{references}

\clearpage
\appendix

\section*{APPENDIX}

\section{Reproducible Implementation Details}
\label{app:reproducible_details}

This appendix provides the operational details needed to reproduce the main TrajWiki pipeline. TrajWiki combines deterministic scoring and validation with structured LLM calls. We therefore distinguish between deterministic candidate construction, structured model decisions, and fallback rules.

\subsection{Memory construction.}
Dialogue messages are first grouped into exchange-level inputs. For each exchange, TrajWiki asks the backbone model to produce an episodic memory seed with three fields:
\texttt{summary\_content}, \texttt{context}, and \texttt{keywords}. A separate claim extraction stage then emits source-grounded atomic claims:
\[
c=(\text{text},\text{status},\text{source\_message\_ids},\text{supporting\_quote}),
\]
where \(\text{status}\in\{\texttt{active},\texttt{deprecated},\texttt{contradictory},\texttt{needs-confirmation}\}\).
All claims are validated against the source messages. Unsupported source ids, empty claims, and ungrounded exact terms are discarded. If the claim extractor fails or omits high-confidence source details, a deterministic preservation fallback extracts must-preserve surfaces such as names, places, books, instruments, activities, counts, temporal expressions, and quoted titles from the raw exchange.

\subsection{Trajectory matching.}
For each new episodic snapshot \(\sigma\), TrajWiki first builds a deterministic feature vector from the snapshot text, extracted claims, and linked source messages. The feature set includes:
\[
K_\sigma,\ E_\sigma,\ X_\sigma,\ F_\sigma ,
\]
where \(K_\sigma\) are lexical keywords, \(E_\sigma\) are entity mentions, \(X_\sigma\) are exact source-backed terms, and \(F_\sigma\) are relation/value facets such as \texttt{home\_country}, \texttt{activity\_location}, \texttt{research\_topic}, and \texttt{event\_type}. For each open trajectory \(T\), TrajWiki stores analogous trajectory-level signals from active claims, historical item terms, retrieval summaries, and the latest snapshot.

Candidate trajectories are prefiltered by keyword/entity/facet overlap. If any overlap exists, only overlapping trajectories are kept, capped at 32 candidates by overlap count; otherwise all open trajectories are allowed as a fallback. Each candidate is scored by
\[
\begin{aligned}
g_{\mathrm{det}}(\sigma,T)
&=0.60\cos(e_\sigma,e_T^{\mathrm{sum}})
+0.20\cos(e_\sigma,e_T^{\mathrm{latest}})
+0.20J(K_\sigma,K_T) \\
&\quad + b_{\mathrm{entity}}
+b_{\mathrm{facet}}
+b_{\mathrm{cont}}
-p_{\mathrm{mismatch}} .
\end{aligned}
\]
Here \(J\) is keyword Jaccard overlap. The entity bonus is \(0.08\) when normalized entity keys overlap. The facet bonuses are \(0.06\) for exact facet-value overlap and \(0.03\) for facet-tag overlap. The continuity bonus is
\[
b_{\mathrm{cont}}=\min(0.14,0.04|\mathrm{specific}(\sigma)\cap\mathrm{specific}(T)|).
\]
A mismatch penalty of \(0.10\) is applied when a candidate shares only a broad entity but no specific continuity terms; a smaller penalty of \(0.04\) is applied when lexical overlap exists without specific continuity.

The top three candidates under \(g_{\mathrm{det}}\) are then passed to a structured \texttt{trajectory\_match} decision:
\[
\texttt{decision}\in\{\texttt{CONTINUE},\texttt{NEW}\},\qquad
\texttt{selected\_candidate}\in\{T1,T2,T3,\texttt{null}\}.
\]
If structured output is unavailable, TrajWiki falls back to a text parser. In mock deterministic mode only, the best candidate is continued when \(g_{\mathrm{det}}\geq \delta\), with \(\delta=0.72\). Thus, \(\delta\) is a deterministic fallback threshold rather than the sole decision rule in remote-model experiments.

\subsection{Claim evolution.}
Within a trajectory, TrajWiki maintains an append-only claim lifecycle. Exact duplicate claims are first matched by normalized text. If the same claim text changes status, TrajWiki emits a status operation: \texttt{DEPRECATE} when the new status is deprecated, otherwise \texttt{REVISE}. For unmatched new claims, previous claims are shortlisted by keyword overlap, keeping at most three candidate previous claims. A structured \texttt{claim\_transition\_judge} call then decides
\[
\texttt{decision}\in\{\texttt{REVISE},\texttt{ADD}\}.
\]
A \texttt{REVISE} operation creates a new claim and marks the selected previous claim as deprecated; an \texttt{ADD} operation appends the claim without replacing prior evidence. If the judge fails or no previous claim candidate exists, TrajWiki falls back to \texttt{ADD}. Model-suggested operations are treated as hints only; the persisted state is derived by the validated deterministic procedure above.

\subsection{Memory Wiki compilation.}
After replaying a sample's dialogue into trajectories, TrajWiki compiles a sample-level Memory Wiki from the current trajectory store. In the implementation used for our experiments, wiki pages are refreshed after memory replay rather than edited one page at a time online. This preserves stable trajectory, snapshot, claim, and source ids while allowing page organization to be recomputed from the latest memory state.

The compiler first creates candidate page seeds:
\[
\eta\in\{\texttt{index},\texttt{entity},\texttt{topic},\texttt{inventory}\}.
\]
The \texttt{index} page links all trajectories. \texttt{entity} pages group trajectories sharing an entity. \texttt{inventory} pages are created for list-like trajectories with multiple exact terms, display items, counts, or inventory-like facets. \texttt{topic} pages group repeated retrieval keywords. Broad entity pages are split into entity-facet pages when they exceed a trajectory threshold. Seeds with more than six trajectories are split into medium-granularity shards using trajectory-summary embeddings; the target page size is four trajectories and the maximum non-index page size is six. Redundant topic seeds are suppressed when at least \(80\%\) of their trajectories are already covered by entity or inventory seeds. A coverage audit then adds rescue pages for any trajectory reachable only through the index page.

Each page is compiled into markdown with required sections: overview, key facts, items/counts, linked trajectories, and conflicts/uncertainty. If the LLM page compiler returns empty text, missing sections, or placeholder-only content, TrajWiki falls back to deterministic markdown generated from linked trajectory summaries and evidence cards. Each page also stores a routing text, keywords, linked trajectory ids, page type, source-backed terms, display items, counts, and a page embedding.

\subsection{Hierarchical retrieval.}
Given a query \(q\), TrajWiki first classifies the query shape using deterministic lexical rules: list-like, count-like, multi-entity, comparison-like, duration-count, and item family. It then embeds the query and routes it to wiki pages. Index pages are suppressed when non-index pages exist, so the wiki directory does not dominate retrieval.

Each page receives dense and sparse scores. The dense score combines page embedding similarity with entity, reflection, granularity, and answer-family adjustments:
\[
s_p^{dense}
=
\cos(e_q,e_p)
+b_{\mathrm{entity}}
+b_{\mathrm{reflect}}
+a_{\mathrm{gran}}
+0.10s_{\mathrm{family}}
-0.12p_{\mathrm{family}} .
\]
The sparse score is lexical overlap over page keywords, title, entity keys, exact terms, source-event terms, and family terms:
\[
s_p^{sparse}
=
J(K_q,K_p)
+0.35s_{\mathrm{family}}
-0.25p_{\mathrm{family}} .
\]
Dense and sparse rankings are fused by reciprocal rank fusion:
\[
\mathrm{RRF}(x)=
\frac{1}{60+r_{\mathrm{dense}}(x)}
+
\frac{1}{60+r_{\mathrm{sparse}}(x)}.
\]
The top page candidate pool is then optionally reranked by the backbone model with a structured textual output. If reranking fails, the RRF order is used.

Selected wiki pages induce a candidate trajectory universe:
\[
\mathcal{C}_q=\bigcup_{p\in\mathcal{P}_q}L_p .
\]
Candidate trajectories are scored similarly:
\[
\begin{aligned}
s_T^{dense}
&=
0.75\cos(e_q,e_T^{sum})
+0.15\cos(e_q,e_T^{latest})
+0.10b_{\mathrm{entity/facet}}\\
&\quad
+0.08s_{\mathrm{family}}
+0.10s_{\mathrm{event}}
-0.04p_{\mathrm{family}},\\
s_T^{sparse}
&=
J(K_q,K_T)
+0.60s_{\mathrm{family}}\\
&\quad
+0.75s_{\mathrm{event}}
-0.25p_{\mathrm{family}} .
\end{aligned}
\]
Dense and sparse trajectory ranks are again fused with RRF. The top trajectory pool is reranked by the backbone model when available.

For list, count, comparison, and multi-entity questions, TrajWiki performs coverage-aware greedy selection after reranking. The first selected trajectory is the strongest relevance anchor. Subsequent selections maximize
\[
\begin{aligned}
S(T)
&=s_{\mathrm{rank}}(T)
+0.90\Delta_{\mathrm{entity}}
+0.65\Delta_{\mathrm{facet}}
+0.30\Delta_{\mathrm{item}}
+0.12\Delta_{\mathrm{term}}\\
&\quad
+0.18\mathbb{I}_{inventory}
+0.18\mathbb{I}_{count}
+0.75s_{\mathrm{family}}
-p_{\mathrm{red}},
\end{aligned}
\]
where \(p_{\mathrm{red}}\) penalizes repeated entities, repeated terms, repeated item terms, repeated clusters, and answer-family mismatch. For non-coverage queries, TrajWiki simply fills from the reranked order.

\paragraph{Evidence expansion and final answer.}
Selected trajectories are expanded into snapshots. TrajWiki first keeps the top snapshot per selected trajectory, then fills the remaining snapshot budget by query-snapshot embedding similarity. The snapshot budget is \(2k\). Depending on the expansion mode, neighboring snapshots and update-linked snapshots are also added. Source messages linked to selected snapshots are compacted into a final prompt context containing wiki pages, selected trajectory summaries, snapshots, active claims, source messages, temporal anchors, and conflict/deprecated-claim diagnostics.

LoCoMo answer generation uses a structured evidence synthesis schema:
\[
\begin{gathered}
\texttt{can\_answer},\quad
\texttt{answer\_type},\\
\texttt{final\_answer},\quad
\texttt{supporting\_facts},\\
\texttt{supporting\_source\_refs},\\
\texttt{counted\_events},\quad
\texttt{excluded\_events},\\
\texttt{uncertainties},\quad
\texttt{abstain\_reason}.
\end{gathered}
\]
All supporting refs must be visible source refs in the retrieved context. Invalid refs are removed. If an answer has no valid source support after validation, TrajWiki converts it to an abstention. Count answers are additionally validated by removing duplicate, future/planned, uncertain, or unsupported events. Type mismatches trigger a typed retry or safe abstention.

\subsection{Retry and repair.}
Retrieval reflection is used only for LoCoMo when the initial answer abstains or retrieval is weak. A retrieval bundle is considered weak if it has no source messages, no candidate trajectories, no selected snapshots, or no active claims in the answer context. Reflection produces a structured query rewrite with target entities, event terms, temporal terms, must-find terms, candidate wiki slugs, and raw search terms. TrajWiki reroutes through the wiki using these hints. If the reflected answer still abstains or the synthesis schema returns \texttt{can\_answer=false}, a raw-message rescue pass retrieves lexical/semantic raw source candidates without changing the memory state.

Post-generation repair is triggered by deterministic checks for unsupported extra list items, unsupported counts, missing supported list items, over-generic items, scope-mismatched items, unresolved bridge aliases, or answer-type mismatch. Repairs are generated using only retrieved evidence. A repaired answer is discarded if it is malformed, loses required supported values, introduces unsupported extras, or fails arbitration; in such cases the initial answer is kept or TrajWiki emits a safe abstention.

\subsection{Hyperparameters.}
Unless otherwise stated, the LoCoMo experiments use \(m=15\) memory trajectories/pages exposed to downstream stages, \(t_{\mathrm{pages}}=15\) wiki pages, \(k=15\) selected trajectories, neighbor radius \(1\), retrieval expansion mode \texttt{update\_linked\_plus\_neighbors}, trajectory candidate prefilter cap \(32\), rerank candidate-pool size \(12\), RRF constant \(60\), maximum non-index wiki page size \(6\), target wiki page size \(4\), and mock-mode trajectory continuation threshold \(\delta=0.72\). The implementation supports remote, local, and OpenAI-compatible providers; structured-output schemas are used when supported, with parser and deterministic fallbacks otherwise.

\section{Evaluation Details}

\subsection{Evaluation Metrics}\label{app:metrics}

We report two automatic text-based metrics, F1 and BLEU-1, following common practice in long-form question answering benchmarks. In our implementation, both metrics are computed over a canonicalized semantic representation rather than raw surface text. This reduces spurious penalties when an answer expresses the correct fact using longer or slightly different natural-language phrasing.

Given a reference answer \(y\) and a model answer \(\hat{y}\), we first extract canonical answer slots from both answers. Let
\[
\mathcal{C}(y)=\{(s_i,v_i)\}_{i=1}^{n},
\qquad
\mathcal{C}(\hat{y})=\{(\hat{s}_j,\hat{v}_j)\}_{j=1}^{m},
\]
where \(s\) denotes a semantic slot identifier and \(v\) denotes a normalized canonical value.

\paragraph{F1.}
We compute F1 using slot-aware semantic matching between reference and candidate canonical values. Unlike exact set overlap, our current implementation uses a conservative soft matching policy within the same slot. A candidate value may match a reference value if they are identical after normalization, differ only by articles, punctuation, or simple inflection, contain benign modifiers such as ``LGBTQ support group'' versus ``support group'', or match through a predefined alias relation such as ``NYC'' and ``New York City''. Contrastive modifiers, numbers, colors, person names, and other distinguishing terms are preserved to avoid over-matching.

Formally, let
\[
\mu\big((\hat{s},\hat{v}),(s,v)\big)\in\{0,1\}
\]
denote whether a candidate slot-value pair matches a reference slot-value pair under this policy. Matching is restricted to pairs with \(\hat{s}=s\), and each candidate value can match at most one reference value. We therefore compute a maximum one-to-one matching
\[
M^\star
=
\arg\max_M
\sum_{((\hat{s},\hat{v}),(s,v))\in M}
\mu\big((\hat{s},\hat{v}),(s,v)\big).
\]
Precision, recall, and F1 are then
\[
P = \frac{|M^\star|}{|\mathcal{C}(\hat{y})|},
\qquad
R = \frac{|M^\star|}{|\mathcal{C}(y)|},
\]
\[
\mathrm{F1} = \frac{2PR}{P+R}.
\]
If both canonical sets are empty, F1 is defined as \(1.0\); if only one side is empty, F1 is \(0.0\). This metric rewards complete semantic coverage of the reference facts, penalizes missing reference facts through recall, and penalizes unsupported extra canonical facts through precision.

\paragraph{BLEU-1.}
We also report BLEU-1 as clipped unigram precision over canonical answer text. We linearize the canonical slot-value representation into a deterministic text sequence, tokenize it after lowercasing and punctuation normalization, and compute
\[
\mathrm{BLEU\text{-}1}
=
\frac{
\sum_{w \in V}
\min\left(\mathrm{count}_{\hat{y}}(w), \mathrm{count}_{y}(w)\right)
}{
\sum_{w \in V}
\mathrm{count}_{\hat{y}}(w)
}.
\]
Unlike standard BLEU, we do not apply a brevity penalty. BLEU-1 therefore measures how much of the candidate's canonical unigram content is supported by the reference answer. If both token sequences are empty, BLEU-1 is defined as \(1.0\); if only one is empty, it is \(0.0\).

Together, F1 evaluates slot-level semantic fact coverage with conservative soft matching, while BLEU-1 measures canonical unigram precision. These deterministic metrics complement the LLM-based judge by providing reproducible lexical-semantic scores over normalized answer content.

\subsection{Offline Counterfactual Retrieval and Context Ablation}
\label{app:offline_ablation}

We provide a more detailed offline ablation analysis to complement the compact results in Table~\ref{tab:offline_ablation_main}. The analysis is conducted on the LoCoMo multi-hop split using the same full TrajWiki run with GPT-4o-mini. The counterfactual settings reuse saved retrieval diagnostics, gold evidence labels, trajectory/page rankings, source references, and token estimates. They therefore evaluate evidence availability and context cost, but do not rerun answer generation. Consequently, we do not report true judge accuracy for counterfactual variants.

\paragraph{Metrics.}
\textbf{Candidate universe} is the average number of memory items considered before final selection. \textbf{Gold Ref Cov.} measures the fraction of gold source references recovered by the selected context. \textbf{Gold Traj. R@15} measures recall of gold trajectories under the top-15 trajectory cutoff. \textbf{All Ref Rate} is the fraction of queries for which all gold source references are recovered. \textbf{Ctx. Tokens} is an offline whitespace-based context token estimate. \textbf{Unsup. Risk} is a proxy for unsupported evidence risk, indicating that selected evidence lacks sufficient source support for the gold references.

\paragraph{Ablation settings.}
\textbf{Full TrajWiki} uses the observed wiki-routed retrieval and source-grounded trajectory expansion. \textbf{Direct Trajectory Retrieval} removes wiki routing and ranks all trajectories in the same dialogue sample. \textbf{Wiki Summaries Only} keeps wiki pages but does not expand them into source-linked trajectory evidence. \textbf{Flat Raw-Memory Retrieval} ranks raw dialogue messages directly using lexical signals. \textbf{Latest Snapshot Only} and \textbf{Latest Two Snapshots} restrict each selected trajectory to its most recent one or two snapshots. \textbf{Source-Linked Claims Only} keeps only claims with explicit source links and is included as a diagnostic proxy rather than a full source-support-constraint ablation.

Table~\ref{tab:offline_ablation_full_32k} reports all variants at the 32K budget. The full TrajWiki setting provides the best balance between gold source-reference coverage, trajectory recall, and candidate-space size. Direct trajectory retrieval considers more than twice as many candidates but recovers less gold evidence under the same offline selection protocol. Wiki-only retrieval has high trajectory recall but zero gold source-reference coverage, showing that wiki summaries alone are not sufficient as answer evidence. Flat raw-memory retrieval is cheap under this lexical selector but has a very large memory universe and high unsupported-evidence risk. The source-linked-claims-only diagnostic has very low context cost but poor coverage, indicating that source filtering alone cannot replace structured retrieval.

\begin{table}[!t]
\centering
\small
\captionsetup{labelfont=bf}
\caption{\textbf{Full offline ablation at 32K context budget.} All settings use \(k=15\). Counterfactual variants report retrieval/context proxies only.}
\label{tab:offline_ablation_full_32k}
\setlength{\tabcolsep}{4pt}
\resizebox{\linewidth}{!}{%
\begin{tabular}{l c c c c c c}
\toprule
Setting
& Cand. Univ.
& Gold Ref Cov.
& Gold Traj. R@15
& All Ref Rate
& Ctx. Tokens
& Unsup. Risk \\
\midrule
Full TrajWiki
& 55.6 & 0.610 & 0.610 & 0.355 & 2.7K & 0.645 \\
Direct Trajectory Retrieval
& 130.4 & 0.356 & 0.346 & 0.128 & 25.6K & 0.872 \\
Wiki Summaries Only
& 41.2 & 0.000 & 0.807 & 0.000 & 25.3K & 1.000 \\
Flat Raw-Memory Retrieval
& 594.7 & 0.237 & 0.457 & 0.082 & 0.7K & 0.918 \\
Latest Snapshot Only
& 15.0 & 0.613 & 0.610 & 0.351 & 6.4K & 0.649 \\
Latest Two Snapshots
& 22.5 & 0.564 & 0.558 & 0.287 & 6.8K & 0.713 \\
Source-Linked Claims Only
& 347.0 & 0.088 & 0.136 & 0.011 & 0.2K & 0.989 \\
\bottomrule
\end{tabular}%
}
\end{table}

Table~\ref{tab:offline_ablation_budget_curve} reports the budget-controlled results for the main settings. Full TrajWiki reaches its evidence coverage with a compact selected context. In contrast, direct trajectory retrieval improves as the budget increases, but even at 32K it remains below full TrajWiki in gold source-reference coverage and trajectory recall. Wiki-only retrieval increases trajectory recall with larger budgets but continues to have zero gold source-reference coverage, because wiki summaries are not source evidence. This supports the need for both hierarchical routing and source-grounded trajectory expansion.

\begin{table}[!t]
\centering
\small
\captionsetup{labelfont=bf}
\caption{\textbf{Budget-controlled offline ablation.} Values are shown for \(k=15\). ``Ctx.'' is estimated context tokens.}
\label{tab:offline_ablation_budget_curve}
\setlength{\tabcolsep}{4pt}
\resizebox{\linewidth}{!}{%
\begin{tabular}{l c c c c}
\toprule
Setting / Budget
& Gold Ref Cov.
& Gold Traj. R@15
& Ctx.
& Unsup. Risk \\
\midrule
Full TrajWiki, 4K
& 0.606 & 0.605 & 2.6K & 0.649 \\
Full TrajWiki, 8K
& 0.610 & 0.610 & 2.7K & 0.645 \\
Full TrajWiki, 16K
& 0.610 & 0.610 & 2.7K & 0.645 \\
Full TrajWiki, 32K
& 0.610 & 0.610 & 2.7K & 0.645 \\
\midrule
Direct Trajectory Retrieval, 4K
& 0.041 & 0.041 & 0.6K & 0.993 \\
Direct Trajectory Retrieval, 8K
& 0.110 & 0.108 & 3.2K & 0.979 \\
Direct Trajectory Retrieval, 16K
& 0.216 & 0.210 & 9.9K & 0.936 \\
Direct Trajectory Retrieval, 32K
& 0.356 & 0.346 & 25.6K & 0.872 \\
\midrule
Wiki Summaries Only, 4K
& 0.000 & 0.276 & 3.3K & 1.000 \\
Wiki Summaries Only, 8K
& 0.000 & 0.442 & 7.1K & 1.000 \\
Wiki Summaries Only, 16K
& 0.000 & 0.656 & 15.1K & 1.000 \\
Wiki Summaries Only, 32K
& 0.000 & 0.807 & 25.3K & 1.000 \\
\midrule
Flat Raw-Memory Retrieval, 4K
& 0.237 & 0.457 & 0.7K & 0.918 \\
Flat Raw-Memory Retrieval, 8K
& 0.237 & 0.457 & 0.7K & 0.918 \\
Flat Raw-Memory Retrieval, 16K
& 0.237 & 0.457 & 0.7K & 0.918 \\
Flat Raw-Memory Retrieval, 32K
& 0.237 & 0.457 & 0.7K & 0.918 \\
\bottomrule
\end{tabular}%
}
\end{table}

\subsection{Computational Cost and Scalability Analysis}
\label{app:cost_analysis}

We provide a detailed cost and scalability analysis for TrajWiki on the LoCoMo multi-hop split using GPT-4o-mini, with \(m=15\), \(t_{\text{pages}}=15\), and \(k=15\). Since TrajWiki uses multiple asynchronous workers, we report accumulated provider-side latency rather than end-to-end wall-clock time. We also separate deployment costs from benchmark-only evaluation costs. This section reports observed TrajWiki costs only; we do not claim a token break-even against proxy baselines without answer-level reruns.

\begin{table}[!t]
\centering
\small
\captionsetup{labelfont=bf}
\caption{\textbf{Phase-level cost breakdown} for TrajWiki on LoCoMo multi-hop. Tokens and latency are aggregated across the full run. Benchmark evaluation is reported separately from deployment cost.}
\label{tab:cost_phase_breakdown}
\setlength{\tabcolsep}{5pt}
\resizebox{\linewidth}{!}{%
\begin{tabular}{l l l r r r}
\toprule
Phase & Deployment Scope & Reuse Scope
& Calls & Tokens & Provider Latency \\
\midrule
Memory construction
& Deployment & Upfront reusable
& 2256 & 23.71M & 25.66h \\
Query-time retrieval
& Deployment & Per query
& 846 & 10.84M & 1.58h \\
Answer generation
& Deployment & Per query
& 564 & 6.92M & 0.44h \\
Repair / validation
& Deployment & Mixed
& 768 & 1.81M & 2.50h \\
Benchmark evaluation
& Benchmark only & Benchmark only
& 846 & 0.42M & 0.45h \\
\bottomrule
\end{tabular}%
}
\end{table}

Table~\ref{tab:cost_phase_breakdown} gives the phase-level cost breakdown. The largest source of token usage is memory construction, which consumes 23.71M tokens, followed by query-time retrieval with 10.84M tokens and answer generation with 6.92M tokens. This confirms that the dominant cost of TrajWiki is not the final answer itself, but the construction and organization of source-grounded long-term memory.

\begin{figure}[!t]
    \centering
    \captionsetup{labelfont=bf}
    \includegraphics[width=0.78\linewidth]{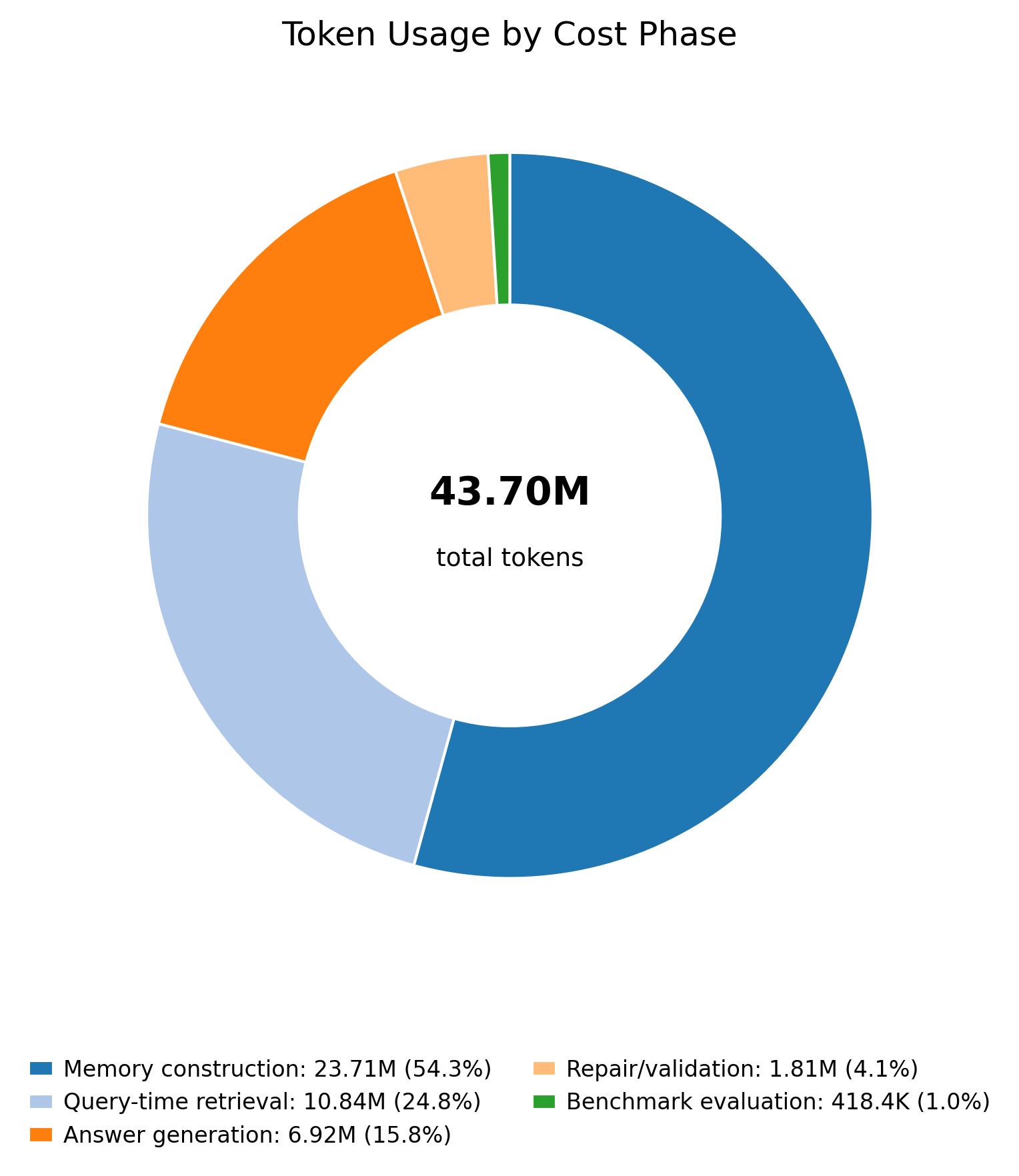}
    \caption{\textbf{Token usage by cost phase.} Token usage is dominated by memory construction and query-time retrieval, while benchmark-only evaluation contributes less than 1\% of total measured tokens.}
    \label{fig:token_usage_by_cost_phase}
\end{figure}

Figure~\ref{fig:token_usage_by_cost_phase} visualizes the same token distribution. Memory construction accounts for 54.3\% of all measured tokens, query-time retrieval accounts for 24.8\%, answer generation accounts for 15.8\%, repair/validation accounts for 4.1\%, and benchmark-only evaluation accounts for only 1.0\%. Thus, evaluation overhead is not the main driver of the reported cost. Instead, the cost reflects TrajWiki's design choice to transform raw dialogue into trajectories, claims, snapshots, and wiki pages before answer generation.

\begin{figure}[!t]
    \centering
    \captionsetup{labelfont=bf}
    \includegraphics[width=0.78\linewidth]{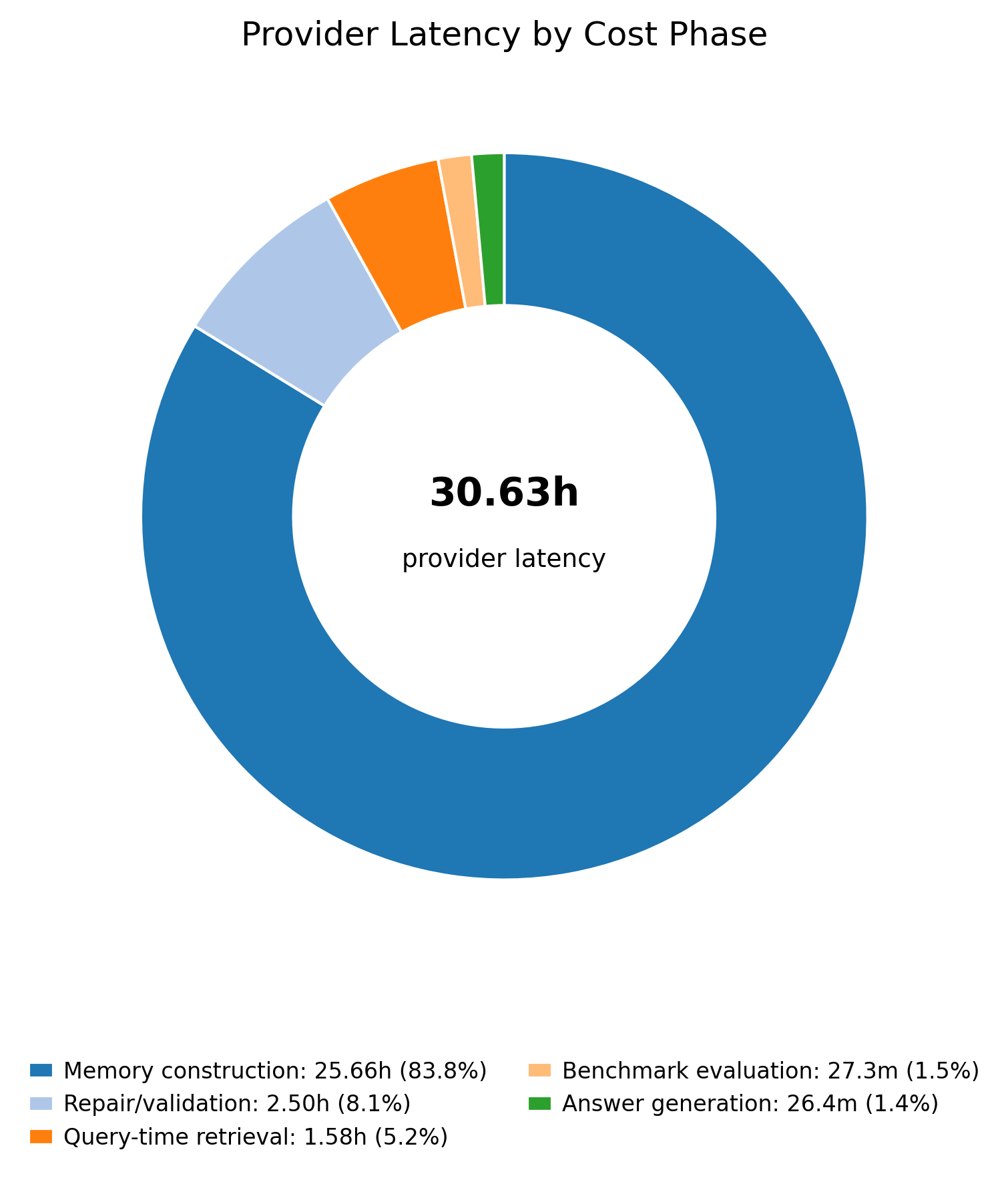}
    \caption{\textbf{Provider-side latency by cost phase.} Accumulated provider latency is primarily spent on memory construction, including extraction, trajectory organization, claim processing, and wiki compilation.}
    \label{fig:runtime_usage_by_cost_phase}
\end{figure}

Figure~\ref{fig:runtime_usage_by_cost_phase} shows an even stronger concentration in provider-side latency. Memory construction accounts for 83.8\% of accumulated provider latency, compared with 8.1\% for repair/validation, 5.2\% for query-time retrieval, 1.5\% for benchmark evaluation, and 1.4\% for answer generation. The latency distribution indicates that memory-building operations such as claim processing, trajectory organization, and wiki construction are the main runtime bottlenecks. This suggests that future efficiency improvements should prioritize caching, batching, and replacing some LLM-based memory-construction steps with cheaper deterministic or distilled components.

\begin{table}[!t]
\centering
\small
\captionsetup{labelfont=bf}
\caption{\textbf{Memory size and candidate scaling.} Values are averaged across LoCoMo dialogue samples or evaluated queries.}
\label{tab:memory_candidate_scaling}
\setlength{\tabcolsep}{5pt}
\resizebox{\linewidth}{!}{%
\begin{tabular}{l r r r}
\toprule
Quantity & Mean & Min & Max \\
\midrule
Raw messages per sample & 588.2 & 369 & 689 \\
Raw memory tokens per sample & 19.9K & 12.0K & 23.9K \\
Trajectories per sample & 128.6 & 43 & 188 \\
Snapshots per sample & 294.3 & 185 & 344 \\
Claims per sample & 3409.2 & 2613 & 4483 \\
Wiki pages per sample & 42.5 & 30 & 59 \\
\midrule
Direct trajectory universe per query & 130.4 & 43 & 188 \\
Wiki-routed trajectory universe per query & 55.6 & 27 & 66 \\
Selected trajectories per query & 15.0 & 15 & 15 \\
Selected snapshots per query & 28.3 & 16 & 30 \\
Selected source messages per query & 74.2 & 29 & 150 \\
\bottomrule
\end{tabular}%
}
\end{table}

\begin{figure}[!t]
    \centering
    \captionsetup{labelfont=bf}
    \includegraphics[width=0.82\linewidth]{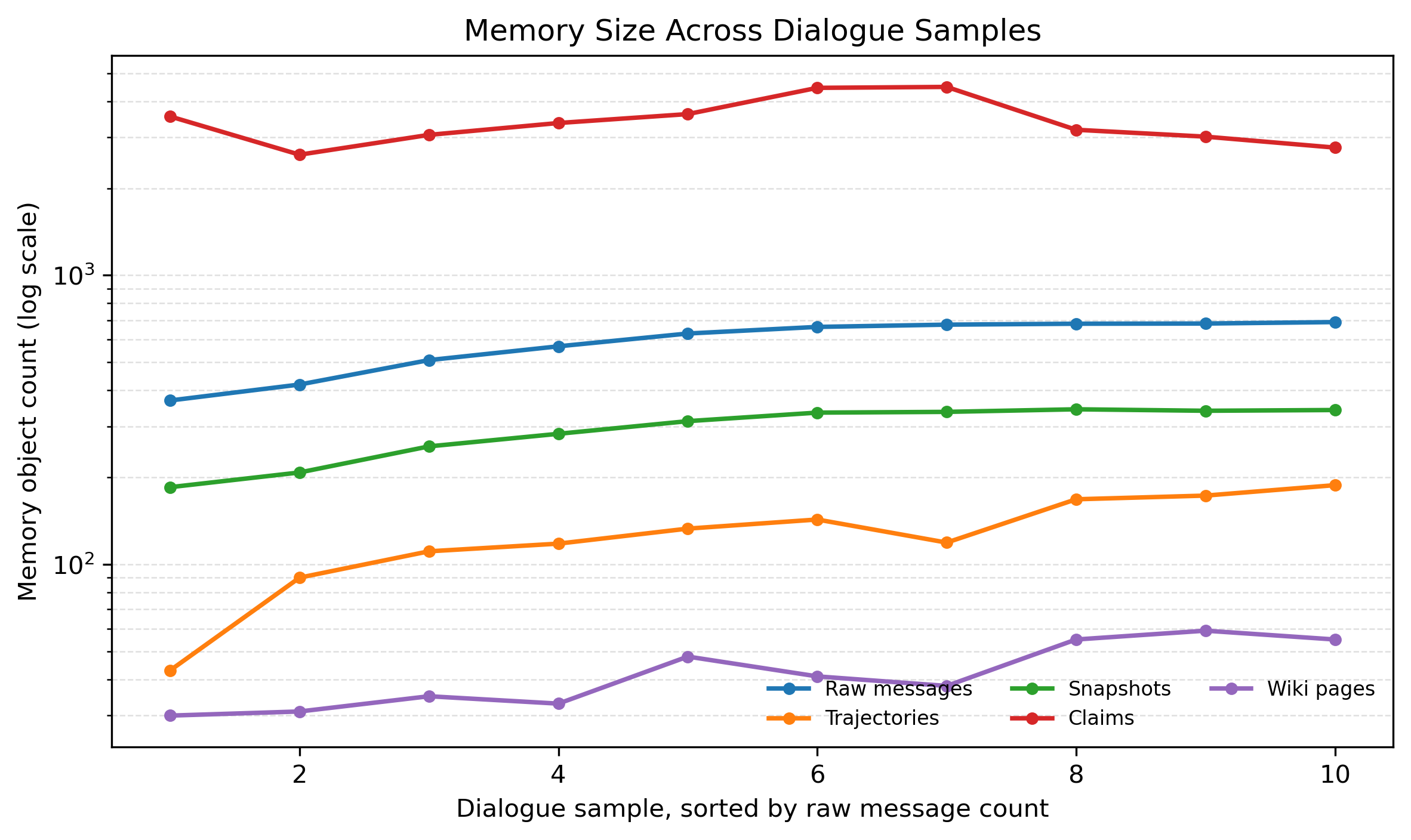}
    \caption{\textbf{Memory size across dialogue samples.} Counts are shown on a log scale. TrajWiki stores raw messages, trajectories, snapshots, claims, and wiki pages as separate memory objects.}
    \label{fig:memory_scaling}
\end{figure}

Table~\ref{tab:memory_candidate_scaling} summarizes how memory size grows across dialogue samples. On average, each dialogue contains 588.2 raw messages, which TrajWiki organizes into 128.6 trajectories, 294.3 snapshots, 3409.2 claims, and 42.5 wiki pages. Figure~\ref{fig:memory_scaling} provides the same view across individual dialogue samples on a log scale. The figure shows that claims are the largest memory object class, while wiki pages remain much smaller in number than trajectories and snapshots. This supports the role of the Memory Wiki as a compact routing layer over a much larger provenance-rich memory store.

\begin{figure}[!t]
    \centering
    \captionsetup{labelfont=bf}
    \includegraphics[width=0.82\linewidth]{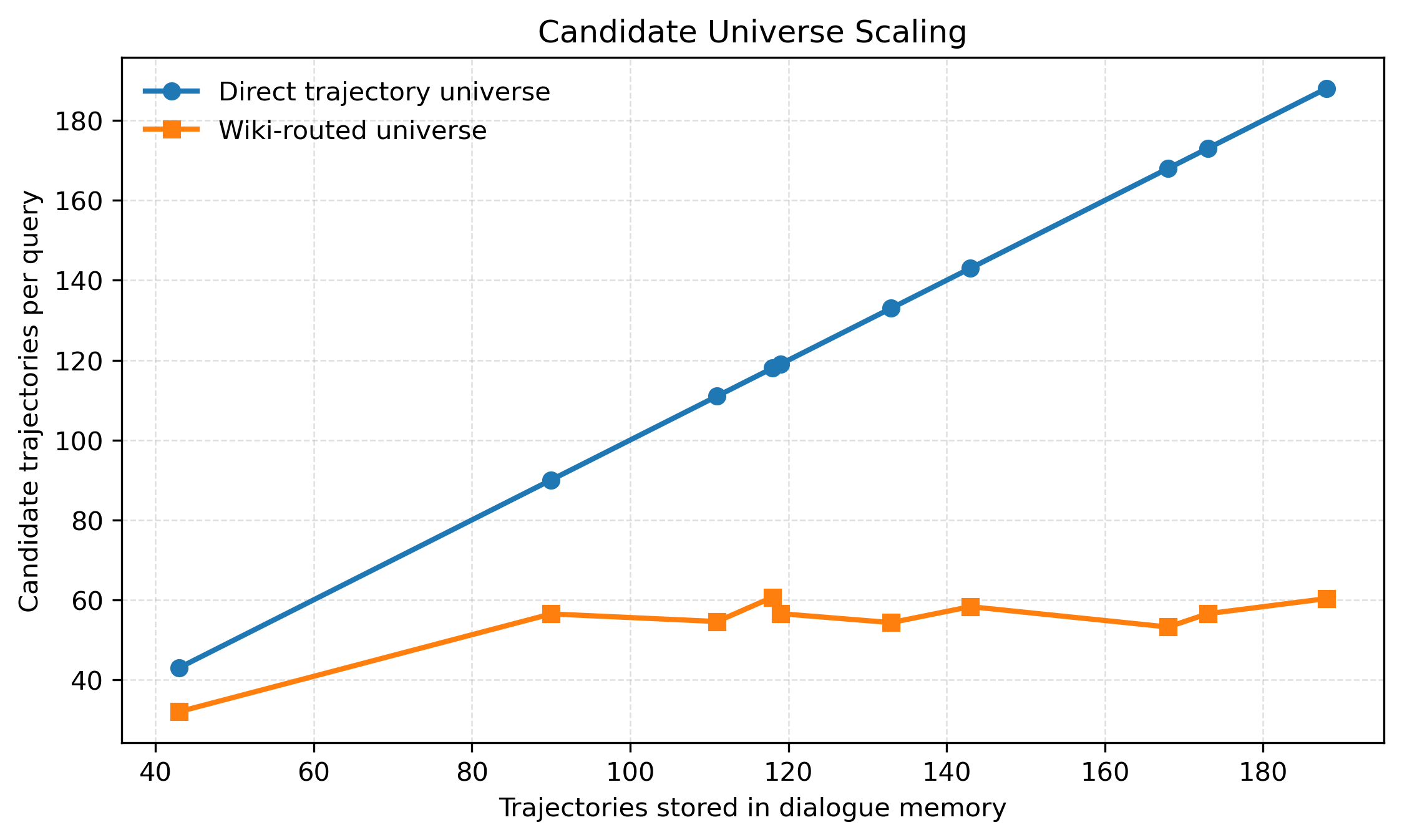}
    \caption{\textbf{Candidate universe scaling.} Wiki routing keeps the trajectory candidate universe substantially smaller than direct retrieval over all stored trajectories.}
    \label{fig:candidate_universe_scaling}
\end{figure}

Figure~\ref{fig:candidate_universe_scaling} shows the query-time effect of this organization. Direct trajectory retrieval would consider 130.4 trajectories per query on average, whereas wiki routing reduces the candidate universe to 55.6 trajectories, a 2.35$\times$ reduction before final top-\(k\) selection. This reduction is important because query-time retrieval otherwise scales with the full trajectory store. The wiki layer therefore provides a concrete scalability benefit even though the current implementation remains token-intensive.

Overall, this analysis shows that TrajWiki is currently a computation-heavy, provenance-oriented memory system rather than a token-minimizing retrieval baseline. Its main efficiency benefit in this implementation is search-space reduction through persistent wiki routing. Future cost optimization should focus on reducing memory-construction latency, caching reusable retrieval computations, and replacing some repeated LLM-based organization steps with cheaper deterministic or distilled alternatives.

\clearpage

\subsection{Limitations and Diagnostic Visibility}
\label{app:limitations}

We provide additional diagnostics for the limitations discussed in Section~\ref{sec:limitations}. These diagnostics are computed offline from the LoCoMo run with GPT-4o-mini. They do not introduce extra LLM calls and should be interpreted as observable proxies rather than human-verified audit accuracy. In particular, we do not report true conflict detection accuracy, obsolete-claim handling accuracy, or human auditor time/error rate without external labels.

\paragraph{Source-support and unsupported-answer diagnostics.}
Table~\ref{tab:auditability_support_diagnostics} summarizes answer-support diagnostics. TrajWiki records supporting references for final answers and compares them with gold source references. The source-supported answer proxy is 0.390, while the unsupported-answer risk proxy is 0.610. The mean retrieved gold-reference coverage is 0.615, but the mean support gold-reference coverage is 0.332, indicating that gold evidence is often retrieved but not always preserved as final answer support. This supports the diagnostic value of TrajWiki: it can distinguish retrieval availability from answer-stage grounding failures.

\begin{table}[!t]
\centering
\footnotesize
\captionsetup{labelfont=bf}
\caption{\textbf{Answer-support diagnostics} on LoCoMo. Metrics are offline proxies computed from recorded support references and gold evidence, not human-verified hallucination labels.}
\label{tab:auditability_support_diagnostics}
\setlength{\tabcolsep}{4pt}
\renewcommand{\arraystretch}{0.97}
\begin{tabular}{l r}
\toprule
Diagnostic & Value \\
\midrule
Evaluated queries & 282 \\
Source-supported answer proxy & 0.390 \\
Unsupported-answer risk proxy & 0.610 \\
Mean retrieved gold-reference coverage & 0.615 \\
Mean support gold-reference coverage & 0.332 \\
Invalid supporting-reference rate & 0.216 \\
Unsupported extra-item count & 30 \\
Unsupported count-error count & 3 \\
Support refs with no gold overlap & 78 \\
\bottomrule
\end{tabular}
\end{table}

\paragraph{Failure localization.}
Table~\ref{tab:failure_localization_diagnostics} reports the distribution of automatically localized failure stages. These stages are derived from recorded provenance, retrieval metadata, gold evidence coverage, and answer-support diagnostics. They are useful for error analysis, but should not be read as failure-localization accuracy because no human stage labels are provided. The largest category is unsupported overgeneration, followed by answer synthesis error. This indicates that many failures occur after relevant evidence is available, rather than only at the retrieval stage.

\begin{table}[!t]
\centering
\footnotesize
\captionsetup{labelfont=bf}
\caption{\textbf{Failure-localization proxy distribution.} The table reports automatically assigned diagnostic stages; no human failure-stage labels are used.}
\label{tab:failure_localization_diagnostics}
\setlength{\tabcolsep}{4pt}
\renewcommand{\arraystretch}{0.97}
\begin{tabular}{l r r}
\toprule
Diagnostic Stage & Queries & Rate \\
\midrule
Unsupported overgeneration & 133 & 0.472 \\
Answer synthesis error & 76 & 0.270 \\
Trajectory selection miss & 28 & 0.099 \\
Page routing miss & 12 & 0.043 \\
Grounding miss & 1 & 0.004 \\
Correct & 32 & 0.113 \\
\bottomrule
\end{tabular}
\end{table}

\paragraph{Independent evaluation and answer-level ablations.}
The original judge-based results have a possible evaluator self-preference confound: GPT-4o-mini and Qwen3-8B judged outputs from their own model settings, while Qwen3-32B was evaluated by the same-family Qwen3-8B model. This concern directly affects judge-based accuracy. F1 and BLEU-1 use the separate canonicalized scoring procedure described in Appendix~B.1.

For the follow-up evaluation, a method-blinded Claude Sonnet 4.6 judge assessed outputs from the full system, ablated variants, and reference systems. This judge was not used for memory construction or answer generation, and variant identities were omitted. This avoids using the generating model or its model family as the judge in the follow-up, but does not remove other automated-evaluation biases or retrospectively change the evaluation of Table~1.

The original counterfactual ablations measured retrieval and context proxies without rerunning answer generation. We therefore reran generation on the same 200 LoCoMo multi-hop queries across the follow-up settings, using GPT-4o-mini, identical answering instructions, and a common 32K-token input-plus-output cap. We report
\begin{equation}
S = 100\,\frac{N_c + 0.5N_p}{N},
\end{equation}
where $N_c$ and $N_p$ count answers judged \texttt{CORRECT} and \texttt{PARTIAL}, respectively. Strict accuracy is $100N_c/N$. Table~\ref{tab:limitations-answer-ablations} reports these answer-level results.

\begin{table}[t]
\centering
\small
\setlength{\tabcolsep}{4pt}
\caption{\textbf{Answer-level ablations on 200 LoCoMo multi-hop queries.} The full and ablated settings use frozen memory stores and a common 32K input-plus-output cap. Score differences are relative to Full TrajWiki, with 95\% confidence intervals from 10,000 paired query-bootstrap draws. Strict accuracy and gold-reference coverage are percentages; input tokens are mean generation inputs, not full-pipeline deployment costs. Source-reference coverage is not reported for the summaries-only condition (N/A).}
\label{tab:limitations-answer-ablations}

\begin{tabularx}{\linewidth}{@{}l r >{\centering\arraybackslash}X r r r@{}}
\toprule
Setting & \shortstack{Judge\\Score} & \shortstack{$\Delta$ vs.\ Full\\{}[95\% CI]} & \shortstack{Strict\\Acc.} & \shortstack{Gold-ref.\\cov.} & \shortstack{Input\\tokens} \\
\midrule
Full TrajWiki        & 42.0 & --- & 20.5 & 60.1 & 19.3K \\
No Wiki Routing      & 31.3 & \mbox{$-10.8\;[-16.0,-5.5]$} & 15.5 & 42.4 & 30.9K \\
Latest Snapshot Only & 37.5 & \mbox{$-4.5\;[-8.5,-0.5]$} & 16.0 & 61.5 & 10.0K \\
Wiki Summaries Only  & 24.8 & \mbox{$-17.3\;[-22.8,-11.8]$} & 10.0 & N/A & 4.5K \\
\bottomrule
\end{tabularx}
\end{table}

The bootstrap intervals quantify query-sampling uncertainty, not variation from repeated stochastic generation. No Wiki Routing bypasses query-time routing while reusing the frozen trajectory store; it is not a reconstruction of the system without Wiki-based organization. Under these conditions, the results support answer-level benefits from Wiki routing and source expansion. Retaining historical snapshots provides a smaller benefit despite comparable aggregate reference coverage, and is not established as necessary for every query.

Simpler systems remain competitive in this follow-up. Matched Full Context achieved a JudgeScore of $39.0$, and the confidence interval for its paired difference from Full TrajWiki included zero. Naive Dense RAG scored $40.3$ with only 1.5K mean input tokens. Thus, these results do not establish universal accuracy or efficiency superiority. 

\paragraph{Computational cost and amortization.}
The observed deployment run used 43.28M tokens, excluding benchmark-only evaluation: 23.71M for memory construction, 10.84M for query-time retrieval, 6.92M for answer generation, and 1.81M for mixed repair and validation. The reported 25.66 hours is accumulated provider-side construction latency, not end-to-end wall-clock time. Reassigning mixed repairs to their corresponding phases yields 24.85M upfront reusable tokens and 18.43M query-side tokens over 282 evaluated queries.

Holding this entire memory collection fixed and using the observed mean query-side cost, amortization over $Q$ queries is
\begin{equation}
\overline{C}_{\mathrm{run}}(Q)
\approx
\frac{24.85\times10^{6}}{Q}
+
\frac{18.43\times10^{6}}{282}
\quad\text{tokens/query}.
\end{equation}
At $Q=282$, the mean is approximately 153.5K tokens per query. The approximately 2.49M upfront tokens per dialogue sample must instead be amortized over the queries served by that individual dialogue, not over the run-wide query count.

This accounting demonstrates reuse under a fixed-memory assumption, not a break-even point against other systems. We did not collect comparable end-to-end cost ledgers for all baselines, including Mem0 and A-MEM, and therefore cannot determine whether the gains justify the additional token or monetary cost. The measured efficiency benefit is a $2.35\times$ reduction in query-time trajectory candidates, from 130.4 to 55.6 on average. It is not evidence of lower total cost.

\paragraph{Conflict, obsolete-claim, and audit-packet diagnostics.}
TrajWiki also records claim lifecycle state and compact audit packets. In this run, conflict-related context was observable for 0.535 of queries, deprecated claims were suppressed in rendered context with rate 1.000, and the deprecated-leakage proxy was 0.128. These numbers show that TrajWiki exposes conflict and obsolete-claim signals, but they are not substitutes for labeled conflict or obsolete-handling accuracy.

The audit packet provides a compact unit for human or post-hoc inspection. Compared with a full-context audit proxy, TrajWiki reduces the mean audit source count from 594.7 to 195.1 and the mean audit packet size from 20.33K to 11.34K estimated tokens. This suggests that the provenance structure can make post-hoc inspection more focused, although actual human audit time and error rate remain future work.

\begin{table}[!t]
\centering
\footnotesize
\captionsetup{labelfont=bf}
\caption{\textbf{Auditability diagnostics.} Conflict and obsolete-claim metrics are observable proxies. Audit-packet size measures the compactness of evidence made available for inspection.}
\label{tab:audit_packet_diagnostics}
\setlength{\tabcolsep}{4pt}
\renewcommand{\arraystretch}{0.97}
\begin{tabular}{l r}
\toprule
Diagnostic & Value \\
\midrule
Conflict exposed rate & 0.535 \\
Deprecated-claim suppression rate & 1.000 \\
Deprecated leakage proxy & 0.128 \\
Obsolete-answer risk proxy & 0.128 \\
\midrule
TrajWiki mean audit source count & 195.1 \\
Full-context proxy source count & 594.7 \\
TrajWiki mean audit packet tokens & 11.34K \\
Full-context proxy audit tokens & 20.33K \\
TrajWiki mean audit claim count & 315.2 \\
\bottomrule
\end{tabular}
\end{table}

\paragraph{Fallback and repair overhead.}
A separate limitation is structured-output dependence. Table~\ref{tab:fallback-repair-risk} summarizes fallback and repair diagnostics. Even with GPT-4o-mini, TrajWiki records substantial repair activity, especially around answer synthesis, validation, and claim preservation. These paths are useful for robustness and diagnosis, but they increase cost and latency. For weaker or non-instruction-tuned models, this dependency is likely to become more severe.

\begin{table}[!t]
\centering
\small
\captionsetup{labelfont=bf}
\caption{\textbf{Fallback and repair diagnostics} on LoCoMo with GPT-4o-mini. Extra tokens and calls quantify the overhead introduced by fallback, repair, and deterministic validation paths.}
\label{tab:fallback-repair-risk}
\setlength{\tabcolsep}{5pt}
\resizebox{\linewidth}{!}{%
\begin{tabular}{l r r r r r}
\toprule
Stage & Events & Fallbacks & Repairs & Risk Events & Extra Tokens \\
\midrule
Answer synthesis / validation & 554 & 0 & 69 & 330 & 25.10M \\
Repair phase & 564 & 0 & 1408 & 158 & 1.84M \\
Semantic metrics & 848 & 1 & 0 & 1 & 0.00M \\
Unknown fallback events & 156 & 7 & 0 & 0 & 0.00M \\
\midrule
Total & 2122 & 8 & 1477 & 489 & 26.93M \\
\bottomrule
\end{tabular}%
}
\end{table}

The ledger reports approximately \(2{,}562\) extra provider calls, 26.93M extra tokens, and 23.90M ms of accumulated provider-side latency from fallback and repair-related paths. These overheads should not be interpreted as end-to-end failures, but they show that reliable structured generation is an operational requirement for TrajWiki.

\paragraph{Case study: summary-mediated salience drift.}
A concrete example is query \texttt{conv-26\_qa\_24}:

\begin{quote}
\textbf{Question:} What does Melanie do to destress? \\
\textbf{Gold answer:} Running, pottery \\
\textbf{Model answer:} Melanie engages in various self-care activities to destress, including running, reading, playing the violin, and prioritizing self-care.
\end{quote}

The raw evidence supports both required items: \texttt{D5:4} states that Melanie signed up for a pottery class and described it as ``therapy for me,'' while \texttt{D7:22} states that she had been running farther to de-stress. This was not a pure retrieval miss: the relevant evidence was present in the memory store. The failure occurred because the pottery fact became less salient through snapshot, trajectory, and wiki-level abstraction. The fact survived in source-linked metadata, but it was not foregrounded strongly enough during final answer synthesis.

This case illustrates the main limitation of provenance-aware summarization. Source links make the failure traceable, but they do not guarantee that every source-specific fact remains salient after repeated abstraction. Future work should combine stricter constrained decoding, periodic source-summary consistency audits, and final-answer verification against raw source evidence rather than generated summaries alone.

\section{Prompts}\label{app:prompts}

This appendix lists the main prompts used by TrajWiki. For tasks using structured output, the prompt is paired with a schema through the provider's structured-output interface; we show the natural-language instruction portion here.

\subsection{Memory Construction Prompts}

\subsubsection*{Episodic Claim Extraction.}
\begin{PromptBox}
TASK=EPISODIC_CLAIM_TEXT_EXTRACT_STRUCTURED
Rewrite the current exchange into readable, source-grounded atomic claim sentences.
Return structured data only through the provider's structured output channel.

Rules:
- Use only facts explicitly stated in the raw messages.
- Resolve speaker names from the raw source lines. If the raw line says "(Melanie): my son", write "Melanie's son", not "Caroline's son", "the user", or "the assistant".
- claim text must be a complete readable factual sentence, not a transcript fragment or broad summary.
- For list-like utterances, include item-level claims for every stated item.
- Preserve exact raw surface phrases for names, titles, places, counts, instruments, events, and items.
- Do not collapse books, instruments, cities, places, activities, events, painted objects, or counts into umbrella-only claims.
- Do not replace concrete raw noun phrases with broader categories: if raw says "sunset", keep "sunset", not only "nature-inspired artwork".
- You may include broader context, but the original source phrase must remain visible in at least one claim.
- supporting_quote must be copied from the raw source message and must support the claim.
\end{PromptBox}

\subsubsection*{Claim Signal Extraction.}
\begin{PromptBox}
TASK=CLAIM_SIGNAL_EXTRACT_STRUCTURED
Extract validated retrieval and display signals from readable claims and source snippets.
Return structured data only through the provider's structured output channel.

Rules:
- Every surface/value/value_span/display value must be copied from a supplied claim or source snippet.
- Do not invent normalized names or semantic replacements.
- Do not output transcript fragments, greetings, acknowledgements, or sentence opener fragments.
- Prefer concise useful retrieval/display signals over exhaustive noisy spans.
\end{PromptBox}

\subsubsection*{Trajectory Matching.}
\begin{PromptBox}
TASK=TRAJECTORY_MATCH_STRUCTURED
Decide whether the new memory continues one of the shortlisted candidate trajectories.
Return structured data only through the provider's structured output channel.

Rules:
- Choose CONTINUE only if the new memory clearly belongs to the same evolving trajectory as one candidate.
- Each candidate includes a trajectory summary for the whole thread and a latest update note for only the newest step.
- Use the trajectory summary as the primary signal; use the latest update note only as recent supporting context.
- Do not require the new memory to resemble only the latest update if the broader trajectory summary clearly matches.
- Same person, broad topic, or generic words such as community/support/art are not enough to CONTINUE.
- CONTINUE requires the same evolving event, project, item, place, count, or fact thread.
- Choose NEW when the memory is about a different item, place, event, count, or topic cluster under the same person unless it explicitly updates the existing thread.
- selected_candidate must be null when decision is NEW.
- selected_candidate must be one of the provided candidate labels when decision is CONTINUE.
- rationale should briefly explain the shared evolving trajectory or why no candidate fits.
- Do not say candidates are merely related or similar.
\end{PromptBox}

\subsection{Memory Wiki and Retrieval Prompts}

\subsubsection*{Trajectory Retrieval Summary.}
\begin{PromptBox}
TASK=TRAJECTORY_RETRIEVAL_SUMMARY
Create a markdown retrieval summary for ONE episodic trajectory using only the supplied grounded internal memory state.

Goal:
- produce a retrieval-oriented synopsis for coarse trajectory selection
- preserve exact names, items, and specific facts
- keep older but query-relevant facts visible even when later snapshots shift topic
- surface uncertainty or conflicts explicitly instead of hiding them

Required headings:
## Profile / Stable Facts
## Item Sets / Named Entities
## Relations / Temporal Updates
## Conflicts / Uncertainty

Rules:
1. Use only the supplied claims, facets, exact terms, and recent snapshot notes.
2. Preserve exact titles, books, recipes, instruments, symbols, places, and named items.
3. Keep list items explicit; do not collapse them into broad themes.
4. Do not let the newest update erase older item sets, places, events, dates, or counts from the trajectory.
5. Do not replace concrete source phrases with broad categories: keep "sunset" visible even if you also mention "artwork".
6. Mention contradictory or uncertain facts under Conflicts / Uncertainty instead of resolving them yourself.
7. Do not add commentary, preamble, or conclusions outside the markdown sections.
8. Exclude filler words, acknowledgements, discourse markers, and casual fragments from the useful retrieval content.
\end{PromptBox}

\subsubsection*{Wiki Page Planning.}
\begin{PromptBox}
TASK=WIKI_PAGE_PLAN
Plan a compact sample-level wiki graph from the provided candidate seed manifest.

Return markdown only with the exact heading:
## Pages

Then emit one bullet per page:
- page_type=<index/entity/topic/inventory> | title=<title> | slug=<slug> | trajectories=<comma-separated trajectory ids> | entities=<comma-separated entities> | links=<comma-separated page slugs>

Rules:
1. Always include exactly one index page.
2. Treat each input block as a candidate seed that already groups related trajectories; refine or select from those seeds instead of inventing broad new groupings.
3. Create entity pages when a person/entity appears across multiple trajectories.
4. Create inventory pages for list/count/item-heavy facts such as activities, books, recipes, instruments, symbols, places, wins, dogs, or repeated events.
5. Create topic pages only when they add navigation value beyond entity or inventory pages.
6. Keep pages compact and specific; avoid broad pages that overlap heavily with existing entity or inventory coverage.
7. Prefer explicit evidence coverage over narrative grouping.
8. Use trajectory historical evidence cards as first-class evidence; do not rely only on a trajectory's latest summary.
9. If a card exposes a specific item, place, event, count, or source anchor, cover it in a non-index page whenever possible.
\end{PromptBox}

\subsubsection*{Wiki Page Compilation.}
\begin{PromptBox}
TASK=WIKI_PAGE_COMPILE
Compile one wiki page from the supplied page metadata, representative trajectory summaries, and historical trajectory evidence cards.

Return markdown only with these headings:
## Overview
## Key Facts
## Items / Counts
## Linked Trajectories
## Conflicts / Uncertainty

Rules:
1. Use only supplied linked trajectory ids, dominant entities, exact terms, facet values, representative trajectory summaries, and historical evidence cards.
2. Keep exact names, items, places, counts, and titles.
3. For inventory pages, favor explicit item lists over prose.
4. Preserve enumerated items and counts explicitly; do not collapse them into broad narrative summaries.
5. Do not replace a concrete source phrase with a broader category: if an evidence card contains "sunset", keep "sunset", not only "nature-inspired artwork".
6. Do not add speculative links or inferred facts not stated in the supplied inputs.
7. In Items / Counts, use only readable display items, readable counts, and readable key facts.
8. If a latest summary and historical card emphasize different facts, present the historical facts in their own concrete bullets instead of dropping them.
9. Do not dump Dominant exact terms verbatim; ignore any low-confidence or fragment-like signal.
10. Do not write placeholder descriptions such as "Not provided", "Unknown", "N/A", "None provided", "No specific key facts", or "No explicit items".
11. Describe a linked trajectory only when its representative summary or evidence card is supplied; otherwise leave the trajectory id without a fabricated description.
12. The system may rewrite the final Linked Trajectories section deterministically, so keep that section concise.
\end{PromptBox}

\subsection{Answer Generation and Evaluation Prompts}

\subsubsection*{LoCoMo Evidence Synthesis.}
\begin{PromptBox}
TASK=LoCoMo_ANSWER_EVIDENCE_SYNTHESIS
Legacy structured LoCoMo evidence synthesis.

Input contains a LoCoMo question followed by retrieved memory/wiki/source evidence.

Rules:
1. Use only retrieved evidence. If it cannot answer the question, set can_answer=false, leave final_answer empty, and explain the gap in abstain_reason.
2. Preserve exact source terms for names, titles, places, dates, counts, activities, and items.
3. supporting_source_refs must use only visible refs such as D8:4, and each ref must support the requested answer family.
4. Count distinct completed events only. Exclude future plans, intentions, reactions, duplicate mentions, general hobbies, and uncertain candidates.
5. Count final_answer must be natural language, not a bare number; use an exact count only when complete evidence supports it, otherwise state a retrieved-evidence lower bound.
6. Date/time answers must use source line date=... fields, snapshot Timestamp lines, and the ## Temporal Anchors block; resolve terms such as "yesterday" to the grounded date when available.
7. For lists/events/preferences, include every in-scope source-backed specific item and avoid scope-mismatched extras.
8. For bridge/alias facts, answer with the concrete source-backed value, e.g. "West County" rather than "old area".

Return structured data only. If structured output is unavailable, return only a JSON object matching the same fields.
\end{PromptBox}

\subsubsection*{LoCoMo Judge.}
\begin{PromptBox}
Evaluate the candidate answer against the question and gold answer.

Equivalence Policy:
- Allow semantic equivalence, near-synonyms, ordering/format variants, harmless modifiers, and concise or longer phrasing with the same meaning.
- Accept date variants for the same time point, e.g. "May 7th" and "7 May"; an unanchored relative date such as "last Tuesday" is PARTIAL, and a clearly wrong date is INCORRECT.
- Treat source-backed aliases as covered: "LGBTQ support group" covers "support group"; "LGBTQ pride parade" covers "pride parade"; "shared her journey at a school event and encouraged students" covers "school speech"; "mentorship program for LGBTQ youth" covers "mentoring program".
- Harmless same-category extras are allowed unless they change the requested scope, contradict the gold answer, or replace a required item.

Strict Cases:
- Counts are strict: "Twice", "2 times", and "two confirmed rejections" are equivalent, but an unqualified wrong count is INCORRECT.
- A retrieved-evidence lower bound such as "The retrieved evidence confirms one rejection" is PARTIAL when the gold requires more.
- Missing a required list item is PARTIAL; for example, omits "school speech" from "pride parade, school speech, support group".
- Wrong entity, polarity, count, time, or place is INCORRECT.

Verdicts:
- CORRECT: all required gold facts/items are covered and no contradiction changes the answer.
- PARTIAL: at least one required fact is covered, but required content is missing or qualified as a lower bound.
- INCORRECT: no required fact matches, or the answer reverses/changes the core meaning.
\end{PromptBox}

\clearpage

\end{document}